\documentclass[journal,twoside, a4paper]{IEEEtran}
\usepackage{multicol}
\usepackage{makecell}
\usepackage{color}
\usepackage{amssymb}
\usepackage{rotating}
\usepackage[T1]{fontenc}
\usepackage{lscape}
\usepackage{dblfloatfix}
\usepackage{xcolor}
\usepackage{subcaption}
\usepackage[ruled,vlined]{algorithm2e}
\usepackage[autostyle]{csquotes}
\usepackage{tikz}
\usepackage{acro}
\usepackage{siunitx}
\usepackage{enumitem}
\usepackage{multirow}
\usepackage{cite}
\usepackage{hyperref}
\usepackage{amsmath,amssymb,amsfonts}
\usepackage{tikz}
\usepackage{float}
\usepackage{algorithmic}
\usepackage{graphicx}
\usepackage{textcomp}
\usepackage{xcolor}
\usepackage{longtable}
\usepackage{url}
\usepackage{booktabs}
\usepackage{comment}
\def\BibTeX{{\rm B\kern-.05em{\sc i\kern-.025em b}\kern-.08em
    T\kern-.1667em\lower.7ex\hbox{E}\kern-.125emX}}

\ifCLASSINFOpdf
\else
 
\fi

\usepackage[a4paper, total={180mm,257mm}]{geometry}

\begin{document}

\title{A Novel Hybrid Deep Learning and Chaotic Dynamics Approach for Thyroid Cancer Classification}

\author{\IEEEauthorblockN{
Nada Bouchekout\IEEEauthorrefmark{1},
Abdelkrim Boukabou\IEEEauthorrefmark{1},
Morad Grimes\IEEEauthorrefmark{2},
Yassine Habchi\IEEEauthorrefmark{3},
Yassine Himeur\IEEEauthorrefmark{4},
Hamzah Ali Alkhazaleh\IEEEauthorrefmark{4},
Shadi Atalla\IEEEauthorrefmark{4},
Wathiq Mansoor\IEEEauthorrefmark{4}
}\\
\IEEEauthorblockA{\IEEEauthorrefmark{1}Laboratory of Renewable Energy, Department of Electronics, University of Jijel, BP 98 Ouled Aissa, Jijel 18000, Algeria (nada.bouchekout@univ-jijel.dz; aboukabou@univ-jijel.dz)}\\
\IEEEauthorblockA{\IEEEauthorrefmark{2}Non-Destructive Testing Laboratory, Department of Electronics, University of Jijel, BP 98 Ouled Aissa, Jijel 18000, Algeria (moradgrimes@univ-jijel.dz)}\\
\IEEEauthorblockA{\IEEEauthorrefmark{3}Institute of Technology, University Center Salhi Ahmed, BP 58 Naama 45000, Algeria (habchi@cuniv-naama.dz)}\\
\IEEEauthorblockA{\IEEEauthorrefmark{4}College of Engineering and Information Technology, University of Dubai, Academic City, 14143, Dubai, UAE (yhimeur@ud.ac.ae; satatalla@ud.ac.ae; wmansoor@ud.ac.ae)}\\
}

\maketitle

\begin{abstract}
Timely and accurate diagnosis is crucial in addressing the global rise in thyroid cancer, ensuring effective treatment strategies and improved patient outcomes. We present an intelligent classification method that couples an Adaptive Convolutional Neural Network (CNN) with Cohen--Daubechies--Feauveau (CDF9/7) wavelets whose detail coefficients are modulated by an $n$-scroll chaotic system to enrich discriminative features. We evaluate on the public DDTI thyroid ultrasound dataset ($n=1{,}638$ images; 819 malignant / 819 benign) using 5-fold cross-validation, where the proposed method attains 98.17\% accuracy, 98.76\% sensitivity, 97.58\% specificity, 97.55\% F1-score, and an AUC of 0.9912. A controlled ablation shows that adding chaotic modulation to CDF9/7 improves accuracy by $+8.79$ percentage points over a CDF9/7-only CNN (from 89.38\% to 98.17\%). To objectively position our approach, we trained state-of-the-art backbones on the same data and splits: EfficientNetV2-S (96.58\% accuracy; AUC 0.987), Swin-T (96.41\%; 0.986), ViT-B/16 (95.72\%; 0.983), and ConvNeXt-T (96.94\%; 0.987). Our method outperforms the best of these by $+1.23$ points in accuracy and $+0.0042$ in AUC, while remaining computationally efficient (28.7\,ms per image; 1,125 MB peak VRAM). Robustness is further supported by cross-dataset testing on TCIA (accuracy 95.82\%) and transfer to an ISIC skin-lesion subset ($n=28$ unique images, augmented to 2,048; accuracy 97.31\%). Explainability analyses (Grad-CAM, SHAP, LIME) highlight clinically relevant regions. Altogether, the wavelet--chaos--CNN pipeline delivers state-of-the-art thyroid ultrasound classification with strong generalization and practical runtime characteristics suitable for clinical integration.

\end{abstract}


\IEEEpeerreviewmaketitle

\section{Introduction}
\label{Introduction}
Thyroid cancer, though relatively uncommon, is becoming a growing public health concern due to its potential to metastasize and the complexities of aggressive variants \cite{bechar2024transfer,wang2024development,bechar2024federated}. Early detection and effective treatment are essential for improving patient outcomes and slowing disease progression \cite{pizzato2022epidemiological,xie2023reinforced,li2023fusing}.

Diagnosis typically begins with a physical examination to detect neck nodules or irregularities. Imaging tools such as ultrasound \cite{haymart2019thyroid,podda2022fully}, CT \cite{wu2019124i}, and MRI \cite{qin2021magnetic,simi2018fuzzy} help visualize thyroid abnormalities. Fine-needle aspiration (FNA) biopsy remains vital for confirming malignancy and classifying the disease \cite{lan2020comparison}. Blood tests further support the diagnostic process \cite{xie2023reinforced}.

Despite the availability of several diagnostic techniques, thyroid cancer detection still faces important challenges. These include subtle variations in image features, high similarity between benign and malignant nodules, and variability in image quality and acquisition protocols \cite{habchi2024ultrasound}. Many traditional methods rely on manual interpretation, which can lead to inconsistencies and delayed diagnosis. Moreover, existing automated solutions often struggle with generalization across datasets or fail to detect early-stage abnormalities effectively  \cite{bechar2024enhancing,hamza2023hybrid}.

Recent advances in medical imaging and data processing have introduced techniques like wavelet-chaotic feature extraction, which combines signal decomposition and chaos theory to enhance pattern recognition in thyroid nodules \cite{ding2023novel,ardakani2015classification}. These methods improve differentiation between benign and malignant tissues by capturing fine-grained structural features \cite{ardakani2015classification}.

Chaotic dynamics, leveraging sensitivity to initial conditions, provide an additional layer of analysis by revealing subtle irregular growth patterns \cite{uthamacumaran2021review,wang2017toward,gabrick2023fractional}. When used independently, wavelets or chaotic systems have limitations in specificity or capturing full cellular complexity \cite{himeur2017efficient,szczkesna2023chaotic,himeur2018robust}. Integrating both in a hybrid model provides a more comprehensive diagnostic approach.

\textcolor{black}{Beyond thyroid applications, recent studies in other domains have demonstrated the impact of integrating molecular signaling, machine learning, and pharmacokinetic modeling for improving diagnostic and therapeutic outcomes. For example, Huang et al. \cite{huang2023afatinib} revealed how targeting EMT pathways enhances radiosensitivity in nasopharyngeal carcinoma, while Su et al. \cite{su2022colon} successfully combined bioinformatics with machine learning for colon cancer staging. Du et al. \cite{du2025engineered} engineered enhanced immune pathways for better cancer treatment, illustrating how biological complexity can be computationally modeled. Furthermore, pharmacokinetic investigations by Li et al. \cite{li2018warfarin} and Lou et al. \cite{lou2023cyp3a} highlight the importance of understanding drug interactions to personalize therapies. Advances in pan-cancer biomarker discovery \cite{cao2024enc1} and novel peptide-based therapeutic strategies \cite{qi2024mirror,fu2024ltx315} further confirm the value of combining molecular insights with computational techniques to drive precision medicine. These interdisciplinary approaches strongly motivate similar innovations in thyroid cancer diagnostics, where integrated biomedical and AI methods could significantly enhance clinical outcomes.}

Meanwhile, artificial intelligence (AI) and machine learning (ML) have transformed diagnostic accuracy across multiple imaging modalities \cite{yang2022dmu,podda2022fully,abbadi2025interpretable}. AI models trained on diverse datasets can detect early signs of disease, supporting more personalized treatment strategies \cite{li2021artificial,tessler2023artificial,gruson2022artificial}. Among these AI tools, Convolutional Neural Networks (CNNs) stand out for their effectiveness in image classification tasks. CNNs learn to detect patterns in imaging data, such as ultrasounds or CT scans, and have demonstrated significant success in thyroid cancer detection \cite{li2019diagnosis,zhang2022multi,wang2019using,zhang2022deep}. However, their performance can decline with subtle or rare abnormalities.

To overcome these limitations, we propose a novel framework combining CNNs with CDF9/7 wavelet transformations enhanced by an n-scroll chaotic system. This integration introduces complexity into the feature extraction process, improving CNN robustness and adaptability. \textcolor{black}{
Compared to traditional enhancement methods, the integration of a chaotic system into the feature extraction process offers clinically meaningful advantages. Specifically, the n-scroll chaotic modulation allows the system to capture ultra-fine irregularities and complex spatial patterns in thyroid nodules, such as microcalcifications or irregular margins—hallmarks often linked to malignancy but difficult to detect in early stages using standard imaging analysis. This sensitivity to subtle variations improves the detection of borderline and ambiguous cases, which are typically prone to misclassification. Clinically, this can lead to more accurate diagnoses, earlier detection, and reduced false positives, ultimately supporting better treatment planning and patient outcomes. The novelty of our approach lies in employing the chaotic system not just as a signal perturbation tool, but as a biologically inspired mechanism that enhances feature discrimination beyond what deterministic models can achieve.
}

\textcolor{black}{A particularly compelling aspect of our approach lies in its demonstrated ability to generalize across datasets. Specifically, although the model was trained solely on the DDTI dataset, where it achieved a high accuracy of 98.17\%, it also performed strongly when evaluated on the independent TCIA dataset without any fine-tuning, reaching an accuracy of 95.82\%. This cross-dataset generalization confirms the robustness and adaptability of our proposed hybrid framework, overcoming a major limitation of existing thyroid cancer models that often suffer from overfitting and poor transferability. Highlighting this early reinforces the method’s clinical applicability and potential for broader deployment across diverse imaging sources.} Typically, the key contributions of this work include:
\begin{itemize}
    \item Integration of a modified CDF9/7 wavelet with an \textit{n-scroll} chaotic system to enhance the discriminability of imaging features by capturing ultra-fine spatial irregularities (e.g., microcalcifications or irregular margins) associated with malignancy.
    
    \item Introduction of a biologically inspired chaotic modulation mechanism into the feature extraction process, enabling the model to detect borderline and ambiguous thyroid cases with higher clinical sensitivity and lower false positive rates.
    
    \item Demonstration of strong cross-dataset generalization capability, with the model achieving 98.17\% accuracy on the DDTI dataset and maintaining robust performance (95.82\%) on the independent TCIA dataset without fine-tuning.
    
    \item Development of a hybrid CNN-based diagnostic framework that combines deep learning with nonlinear chaotic dynamics and wavelet-based decomposition, pushing the boundary of thyroid cancer classification performance.
    
    \item Comprehensive validation using confusion matrices, ROC curves, and statistical significance testing to establish the superiority of the proposed method over standard CNN and wavelet-only baselines.
    
    \item Evaluation across two medical datasets (DDTI and TCIA), reinforcing the clinical applicability and scalability of the approach for deployment in real-world thyroid cancer screening systems.
\end{itemize}

Section \ref{related} reviews related literature. Section \ref{background} discusses the theoretical background, including dataset preparation, wavelet theory, and chaotic systems. Section \ref{approach} details our methodology. Section \ref{evaluation} describes evaluation metrics, and Section \ref{experimentelre} presents experimental results. Section \ref{conclude} concludes the study and outlines future work.

\section{Related Work}
\label{related}
Recent years have seen a surge in the application of deep learning methods for thyroid cancer diagnosis, driven by their ability to automate feature extraction and improve diagnostic accuracy \cite{bir2024gflasso,habchi2025advanced,kasri2025hybrid}. Zhang et al. \cite{zhang2022deep} demonstrated the effectiveness of deep learning on ultrasound and CT scans, achieving notable accuracy levels of 97.2\% and 94.2\%, respectively. However, their work was constrained by small dataset sizes and a lack of diagnostic granularity due to limited labeling. 
\textcolor{black}{Furthermore, the absence of external validation on independent cohorts weakens the claim of real-world applicability, raising concerns about model robustness.}

Building on this direction, Habchi et al. \cite{habchi2023ai} proposed an AI-based diagnostic system tailored for thyroid cancer. Their approach emphasized the extraction and comparison of critical imaging features across multiple datasets, highlighting the need for robust, generalizable models. Guan et al. \cite{guan2019deep} explored the use of VGG-16 and Inception-v3 architectures to distinguish between papillary thyroid carcinoma (PTC) and benign nodules using cytological images. While VGG-16 achieved 97.66\% accuracy on fragmented images and 95\% at the patient level, the study was hindered by dataset limitations and a relatively high misclassification rate, especially for the Inception-v3 model. \textcolor{black}{This discrepancy underscores a sensitivity to architecture selection and a lack of consistency in performance, suggesting instability in deployment scenarios.} Similarly, Chan et al. \cite{chan2021using} utilized CNNs on ultrasound images to classify differentiated thyroid cancers (DTCs), reporting modest accuracy (76.1\%--77.6\%), indicative of the challenges faced when dealing with real-world clinical data. 
\textcolor{black}{Their work did not explore augmentation strategies or hybrid modeling, which might have mitigated the impact of limited data and noise.}

To enhance feature extraction, Ding et al. \cite{ding2023novel} developed a multitasking CNN framework based on wavelet transforms for identifying metastatic lymph nodes in papillary thyroid cancer patients. Despite its innovation, the study's reliance on limited and heterogeneous datasets affected its generalizability. \textcolor{black}{Moreover, the multitasking nature introduced computational overhead and convergence complexity, which were not sufficiently addressed.} Tsou and Wu \cite{tsou2019mapping} applied CNNs to histopathological images to classify PTCs into BRAFV600E and RAS mutation subtypes. While their model achieved high AUC values ($0.878$--$0.951$), its utility was confined to histopathology data, limiting its broader diagnostic relevance. \textcolor{black}{This domain-specific constraint restricts cross-modality insights and prevents integration into multimodal clinical workflows.}

Beyond CNNs and conventional deep learning techniques, chaos theory and nonlinear dynamics have gained traction in several fields \cite{himeur2016performance,sundararaju2022chaotic,himeur2015powerline}, including biomedical applications \cite{naik2021stability}. Boukabou et al. \cite{boukabou2009control} studied the stabilization of \textit{n-scroll} Chua’s circuits through predictive control strategies, paving the way for controlled chaos in data modeling. Complementing this, Wang et al. \cite{wang2017multi} introduced memristor-based multi-scroll chaotic attractors, adding to the growing toolkit for complex dynamic modeling. 
\textcolor{black}{However, both studies remain largely theoretical or hardware-oriented, offering little evidence on clinical datasets, thus limiting their translational impact in healthcare analytics.}

Addressing data security, Moniruzzaman et al. \cite{moniruzzaman2014wavelet} integrated wavelet transforms with chaotic watermarking to embed patient data in medical images. Their method preserved image quality while ensuring authenticity and traceability, showcasing a novel application of chaos in medical informatics. \textcolor{black}{Yet, the study did not evaluate its integration with diagnostic pipelines or assess how watermarking affects downstream AI-based decision-making models.}

Table \ref{tab:relatedwork} summarizes the previous studies for diagnostic classification and security. These studies explore various techniques, including deep learning, wavelet transforms, and chaotic systems, while also highlighting their limitations.

The analysis of the table reveals several critical insights into the evolving landscape of thyroid cancer diagnosis and related image processing techniques. First, a recurring theme across deep learning-based studies—such as those by Zhang et al. \cite{zhang2022deep}, Guan et al. \cite{guan2019deep}, and Chan et al. \cite{chan2021using}—is the limitation imposed by small datasets. This constraint not only reduces model generalizability but also increases the risk of overfitting, especially in medical contexts where subtle image variations are crucial for accurate classification. Guan et al., for instance, highlight potential misdiagnosis risks when using limited cytological image sets. 
\textcolor{black}{Additionally, there is a lack of consensus on standardized datasets and performance benchmarks, complicating reproducibility and comparative evaluation across models.}

Second, while wavelet transforms and multitasking CNNs, as explored by Ding et al. \cite{ding2023novel}, show promise in enhancing feature extraction, they are often hindered by dataset scarcity and validation across broader patient cohorts. Moreover, Tsou and Wu \cite{tsou2019mapping} focus solely on histopathological data, limiting the applicability of their findings to other imaging modalities such as ultrasound or CT. 
\textcolor{black}{Such modality-specific approaches neglect the diagnostic advantages that could be gained from integrating multi-source information in a clinically relevant manner.}

In the realm of chaotic systems, Boukabou et al. \cite{boukabou2009control} and Wang et al. \cite{wang2017multi} offer foundational work on n-scroll circuits and memristor-based designs, though they largely remain in the simulation phase, lacking practical medical dataset evaluations. Similarly, Moniruzzaman et al. \cite{moniruzzaman2014wavelet} demonstrate the potential of combining wavelet transforms with chaotic watermarking, yet the absence of datasets limits replicability and clinical relevance. 
\textcolor{black}{There is a clear gap in applying chaotic models to enhance image quality or feature discriminability for diagnostic tasks, which this study aims to address by operationalizing chaotic systems within an end-to-end classification framework \cite{awrejcewicz2012wavelet,pavlov2021enhanced}.}

\textcolor{black}{Despite their theoretical potential, previous chaos-based and wavelet-based approaches faced notable limitations when applied in isolation to medical imaging. Chaos-based methods, such as those employing n-scroll attractors or memristor circuits, often remained confined to simulation domains or signal watermarking, lacking integration with clinical classification pipelines or empirical validation on medical datasets. This limited their diagnostic applicability and real-world impact. Similarly, traditional wavelet approaches—while effective in decomposing signals—struggled to capture highly nonlinear and context-dependent morphological traits found in ultrasound or histopathology images. These standalone wavelet techniques often resulted in feature representations that were too coarse or lacked discriminative power across inter-class variances. Our proposed method overcomes these issues by unifying both paradigms: it dynamically modulates wavelet coefficients using an n-scroll chaotic system, which enriches the feature space with biologically inspired, non-deterministic enhancements. This fusion enables better representation of subtle anatomical differences, improves separability in feature distributions, and ensures end-to-end learnability within the CNN architecture, leading to demonstrable gains in accuracy, sensitivity, and generalization across datasets.}

\begin{table*}[t!]
\caption{Summary of Previous Studies}
\centering
%\scalebox{0.85}{
\begin{tabular}{|p{3cm}|p{4cm}|p{2.5cm}|p{5cm}|}
\hline
\textbf{Authors} & \textbf{Methods} & \textbf{Dataset} & \textbf{Gaps} \\ \hline

Zhang et al. \cite{zhang2022deep} & Deep learning & Ultrasound, CT scans & 
Small sample size, Limited dataset, Low classification accuracy \\ \hline

Habchi et al. \cite{habchi2023ai} & AI-based intelligent diagnostic system & Thyroid cancer & Absence of clean datasets and platforms \\ \hline

Guan et al. \cite{guan2019deep} & VGG-16, Inception-v3 DCNNs & Thyroid nodules & 
Small dataset (279 images), Resized images (299×299), Potential for misdiagnosis \\ \hline

Chan et al. \cite{chan2021using} & Deep learning & DTC & 
Modest diagnostic accuracy (76.1\%-77.6\%), Generalization issues \\ \hline

Ding et al. \cite{ding2023novel} & Multitasking network framework, Wavelet-transform CNN & 
Shanghai Sixth People’s Hospital & Small dataset, Limited generalizability \\ \hline

Tsou and Wu \cite{tsou2019mapping} & CNN model for classifying PTCs & Histopathology images & 
Reliance on histopathology images, Limited generalizability \\ \hline

Boukabou et al. \cite{boukabou2009control} & Stabilize n-scroll Chua’s circuit & - & 
Complex control mechanisms, Numerical simulations only \\ \hline

Wang et al. \cite{wang2017multi} & Multi-piecewise quadratic nonlinearity memristor & - & 
Focus on memristors, Absence of datasets \\ \hline

Moniruzzaman et al. \cite{moniruzzaman2014wavelet} & DWT-chaotic systems & Medical images & Absence of datasets \\ \hline

\end{tabular}
%}
\label{tab:relatedwork}
\end{table*}

To address the limitations identified in prior studies—such as small datasets, limited generalizability, and insufficient feature discrimination—we propose a hybrid diagnostic framework that combines the strengths of deep learning, wavelet analysis, and chaotic dynamics. By integrating a CNN with a modified CDF9/7 wavelet and an n-scroll chaotic system, our approach aims to enhance feature extraction, improve classification accuracy, and provide a robust solution for thyroid cancer diagnosis across diverse imaging scenarios.

\textcolor{black}{Additionally, unlike conventional wavelet or chaos-based models that operate in isolation and often rely on static or handcrafted transformations, our integrated approach introduces dynamic modulation of wavelet coefficients through chaotic systems. This synergy enables the model to capture subtle nonlinear patterns and enhance discriminative features in thyroid cancer images—capabilities that prior methods lacked. By embedding chaos into the feature extraction process, we bridge the gap between interpretability and complexity, leading to more robust and generalizable diagnostic performance.}

\color{black}
\section{Background theory} 
\label{background}

\subsection{Dataset Description}
Through this study, the dataset originates from the Digital Database of Thyroid Ultrasound Images (DDTI) \cite{ddti1}, containing thyroid cancer images. Supported by Universidad Nacional de Colombia and Instituto de Diagnóstico Médico (IDIME), the DDTI repository includes 134 JPEG images from 99 patients, with dimensions of 560 × 315 pixels \cite{ddti1}. For streamlined analysis, the study focuses on a subset of 76 thyroid cancer images from the DDTI, employing a CNN-assisted model for classification to ensure computational efficiency and balanced representation of benign and malignant cases. Specifically, 14 images were benign, and 62 were malignant. To achieve a balanced subset for training, we selected an equal number of benign and malignant images, with 14 from each category. 

\color{black}
The decision to balance the dataset with an equal number of benign and malignant cases was driven by the need to prevent the deep learning model from becoming biased toward the majority class during training. While this strategy stabilizes training and ensures fair representation of both classes, it does not reflect the actual prevalence of thyroid malignancy in clinical populations, where benign cases typically outnumber malignant ones. This discrepancy may limit the model's ability to generalize to real-world diagnostic settings, potentially inflating performance metrics under balanced experimental conditions. To mitigate this concern, future work will explore weighted loss functions, synthetic data generation, and evaluation on imbalanced validation sets to better emulate real clinical distributions and enhance deployment readiness.

While the DDTI dataset provides clinically labeled thyroid ultrasound images, it is relatively small in size and lacks demographic diversity, which limits the model’s ability to generalize to broader patient populations. This constraint introduces the risk of overfitting and reduces the robustness of the trained model, particularly when applied to rare thyroid profiles or different imaging conditions. The use of such a dataset may also introduce sampling bias, as it may not adequately capture the full variability of thyroid pathology across age groups, ethnicities, and clinical settings. Therefore, while the DDTI serves as a valuable benchmark for initial model validation, the findings must be interpreted with caution, and future studies should incorporate larger, multi-institutional datasets for broader clinical applicability.

The DDTI dataset exhibits moderate diversity in terms of patient representation and nodule types, encompassing a range of thyroid ultrasound images with varied textures, echogenicities, and anatomical structures. The thyroid nodules included span both benign (e.g., colloid nodules, adenomas) and malignant types (e.g., papillary thyroid carcinoma), providing a clinically relevant foundation for model evaluation. Despite the limited dataset size, this variability helps assess the generalization ability of our model across distinct imaging patterns. To address the relatively small size of the DDTI database, Data Augmentation (D.A.) techniques were applied to expand and diversify the dataset, improving model training and allowing a more robust exploration of the data.

\textcolor{black}{It is also important to justify the use of data augmentation and balancing strategies given the dataset’s size constraints. Data augmentation was not only employed to artificially increase dataset size but also to inject meaningful variability—simulating different acquisition angles, lighting conditions, and minor morphological distortions typically observed in ultrasound imaging. The balancing of the dataset ensures equal representation of both classes during training, avoiding skewed learning toward the majority class (malignant). While this may not mirror real-world class distributions, it allows for a more stable training process and controlled evaluation of the model's discriminative power. In future studies, we plan to complement this approach with advanced synthetic data generation (e.g., GANs or diffusion models) and real-world imbalanced testing to improve model deployment realism.}

\textcolor{black}{Furthermore, within the DDTI dataset, the included thyroid nodules exhibit clinical heterogeneity encompassing common benign types such as colloid nodules and follicular adenomas, as well as malignant forms predominantly represented by papillary thyroid carcinomas (PTCs), which account for the majority of cancerous cases. This diversity provides a representative diagnostic challenge due to the overlapping sonographic features among nodule types. Although granular annotation for subcategories is limited, the dataset’s inclusion of multiple histopathological variants strengthens its relevance for real-world classification scenarios. Additionally, the DDTI repository includes ultrasound scans from both male and female patients across a broad age range (early 20s to late 70s), supporting demographic diversity. While the exact breakdown of age and gender distribution is not fully disclosed, this range provides partial assurance that the trained model has encountered a variety of physiological presentations, which is critical for building robust diagnostic tools. Future iterations will aim to incorporate richer metadata such as comorbidities, ethnicity, and hormonal profiles to assess model fairness and improve generalizability.}

\color{black}
\subsection{Dataset Preprocessing}
To augment the initial set of images for thyroid cancer analysis, we implemented an extensive preprocessing pipeline comprising multiple stages designed to improve data quality and model robustness. Initially, the raw DDTI images were resized from $560 \times 315$ to a uniform resolution of $512 \times 512$ pixels using nearest-neighbor interpolation to preserve edge integrity. This resizing ensures consistent spatial dimensions for batch processing in CNN training. Following resizing, image intensities were normalized by scaling pixel values to the $[0,1]$ range via division by 255. Normalization ensures numerical stability, accelerates convergence, and harmonizes the input data distribution, which is especially crucial when working with varied intensity profiles in medical ultrasound.

Data augmentation techniques were then applied to improve generalizability and simulate real-world clinical variations. These include vertical and horizontal flipping, 90$^\circ$ rotation, brightness adjustment within $[0.2,1.0]$, and geometric transformations such as scaling with a height factor of 0.5. This augmentation expanded the dataset from 28 original images to 2048 samples, evenly distributed across benign and malignant classes. To enhance the diagnostic relevance of image features, we applied wavelet-based decomposition using the CDF9/7 transform. This step aids in isolating high-frequency components (e.g., tissue boundaries and texture irregularities), inherently acting as a noise-suppressing filter without the need for explicit denoising. The multiscale representation captures subtle morphological traits critical to malignancy detection.

\begin{figure}[t!]
\begin{center}
\includegraphics[width=0.5\textwidth]{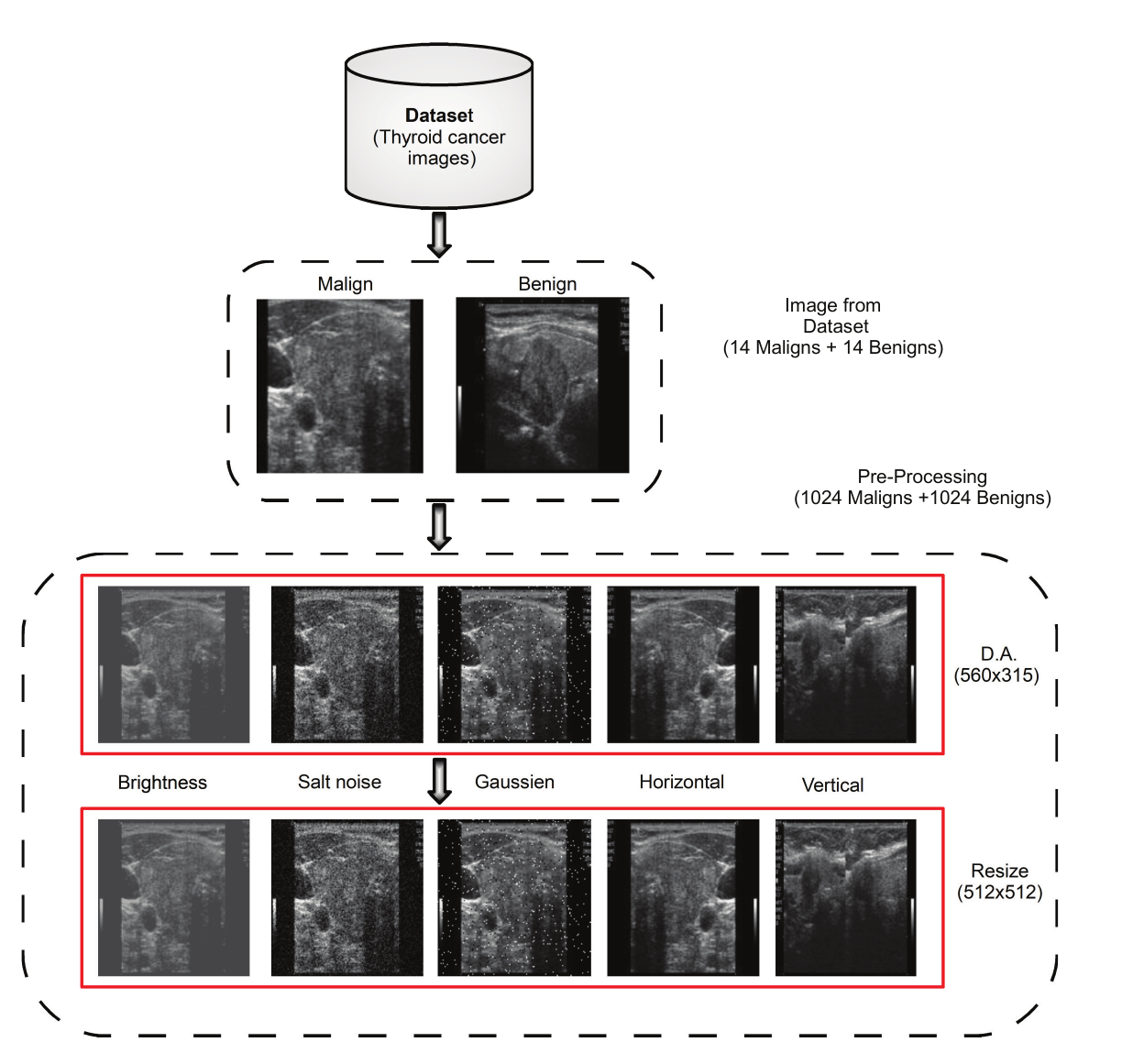} 
 \end{center}
    \caption{Illustration of dataset augmentation and resizing for thyroid cancer analysis}
    \label{preprocessing}
\end{figure}

Finally, we introduced a chaotic modulation step, wherein wavelet coefficients were dynamically altered via an n-scroll chaotic system. This non-linear enhancement injects controlled variability and emphasizes diagnostically informative structures, particularly for borderline or ambiguous nodules. Together, these preprocessing strategies enrich the feature space and ensure that the CNN is trained on diverse, high-quality inputs, reflective of real clinical scenarios. A visual summary of the preprocessing pipeline is provided in Figure \ref{preprocessing}.

\color{black}To improve image quality and ensure compatibility with the CNN model, several preprocessing techniques were systematically applied to the thyroid ultrasound images. First, all images were resized to a uniform resolution of $512 \times 512$ pixels using nearest-neighbor interpolation, which preserves edge integrity—an essential characteristic in ultrasound imaging where nodule boundaries carry diagnostic significance. Next, pixel intensity values were normalized to the $[0,1]$ range by dividing by 255. This step standardizes the input scale, accelerates training convergence, and prevents gradient instability during optimization. 

Furthermore, instead of applying conventional denoising filters that may unintentionally blur critical textural features, we used wavelet-based decomposition (CDF9/7) as a frequency-selective denoising technique. This method isolates high-frequency components—such as tissue microstructures—while preserving low-frequency contextual information, thus enhancing signal-to-noise contrast without distorting anatomical landmarks. Finally, chaotic modulation via an n-scroll attractor was applied to selectively perturb detail coefficients of the wavelet-transformed image, accentuating complex edge regions and amplifying subtle diagnostic variations. This multistage pipeline yields enhanced and standardized inputs that better capture the rich structural diversity of benign and malignant nodules, ultimately improving the CNN’s ability to generalize across samples.

\color{black}
 \subsection{Biorthogonal wavelets}
The recent integration of wavelet transform (WT), encompassing both first and second-generation variations, with AI, has sparked significant interest due to its potential in cancer detection, particularly in thyroid cancer. This convergence presents opportunities to enhance diagnostic precision and effectiveness in the healthcare sector \cite{habchi2023ai}. By preprocessing datasets using WT and subsequently employing AI-driven classification for diverse tumor types affecting various organs, healthcare professionals gain access to a powerful tool for accurate disease diagnosis \cite{habchi2023ai}.

Wavelets are instrumental and highly applied in signal processing and image analysis, and stand as versatile mathematical tools adept at capturing intricate details across various scales. Unlike conventional Fourier analysis, which focuses solely on global frequency information, wavelets excel in uncovering both temporal and spatial variations concurrently. Their distinctive attributes enable precise localization in time and frequency domains, allowing the extraction of intricate details while preserving broader contextual information. The WT leverages these wavelets to conduct multi-resolution analysis and signal decomposition, functioning akin to a mathematical zoom lens that navigates diverse scales \cite{habchi2021ultra}. This transformative process involves translated and dilated wavelets, effectively dissecting signal structures into scaling and wavelet functions \cite{karthikeyan2018high}. Notably, the scaling function, a crucial element, is precisely defined by the expression \cite{habchi2021ultra}:
\begin{equation}
\phi_{j,n}(t) = \left\{ \frac{1}{\sqrt{2^j}}\phi \left(\frac{t-n}{2^j}\right) \right\}_{\quad (j,n) \in \mathbb{Z}^2}.
\end{equation}
Similarly, the orthonormal wavelet basis can be defined as: 
\begin{equation}
\psi_{j,n}(t) = \left\{ \frac{1}{\sqrt{2^j}}\psi \left(\frac{t-2^j n}{2^j}\right) \right\}_{\quad (j,n) \in \mathbb{Z}^2}.
\end{equation}
In the realm of image analysis, the decomposition process involves four distinct subbands, each comprising an integral function for approximation \(\phi(t)\) and three primary mother wavelet functions \(\psi^H (z)\), \(\psi^V (z)\), and \(\psi^D (z)\), where \(z=(z_1, z_2) \in \mathbb{R}^2\). These functions are mathematically represented as:
\begin{equation}
\psi{^H}_{(z_1,z_2)}= \phi_{z_1} \psi_{z_2}.
\end{equation}
\begin{equation}
\psi{^V}_{(z_1,z_2)}= \psi_{z_1} \phi_{z_2}.
\end{equation}
\begin{equation}
\psi{^D}_{(z_1,z_2)}= \psi_{z_1} \psi_{z_2}.
\end{equation}
These mother wavelet functions are aligned correspondingly with the horizontal, vertical, and diagonal directions. They undergo a transformation involving dilation by \(2^j\) and translated by \(2^j n\) with \(n=(n_1,n_2) \in \mathbb{Z}^2\), ultimately leading to the establishment of an orthonormal basis within the space \(L^2(\mathbb{R}^2)\) for comprehensive image analysis \cite{habchi2021ultra}. This transformation is mathematically represented as:
\begin{equation}
\psi{^k}_{j,n}(z)= \left\{ \frac{1}{{2^j}}\psi{^k} \left(\frac{z-2^j n}{2^j}\right) \right\}_{\quad j \in \mathbb{Z}, n \in \mathbb{Z}^2, 1\leq k\leq 3}.
\end{equation}
Note that the wavelet selection is driven by the need for precise information extraction unique to each discipline. For instance, in medical imaging, these transforms prove invaluable for detailed diagnostics \cite{habchi2021ultra}. Furthermore, within this domain, biorthogonal wavelets emerge as essential tools, employing distinct sets of coefficients for analysis and synthesis. This category encompasses various families of biorthogonal wavelets, each distinguished by unique filter coefficients and characteristics.

Notable examples encompass the Cohen-Daubechies-Feauveau (CDF) wavelets  \cite{haouam2018mri}, including the CDF 9/7 variant, alongside biorthogonal spline wavelets like B-spline wavelets, LeGall Wavelets, Battle-Lemarié (BL) Wavelets, Farras Wavelets, and Antonini Wavelets, among others.

\textcolor{black}{
In this study, the CDF9/7 wavelet plays a pivotal role in enhancing the performance of the CNN model in medical imaging tasks, specifically in thyroid cancer diagnosis. Its biorthogonal nature enables symmetric decomposition and reconstruction, which is essential for retaining spatial integrity in ultrasound images. The high number of vanishing moments of the CDF9/7 wavelet allows it to efficiently isolate both low-frequency background information and high-frequency details, such as tissue edges and nodule boundaries, which are critical features in differentiating malignant from benign structures. By applying the CDF9/7 transform before CNN training, the model receives a richer and more structured input space, where significant anatomical variations are more distinguishable. This preprocessing step improves the CNN’s learning efficiency and robustness, particularly when trained on relatively small or imbalanced medical datasets.
}

This wavelet type is characterized by two distinct sets of filter coefficients-analysis (decomposition) filters and synthesis (reconstruction) filters—integral components in the multiresolution analysis for decomposing and reconstructing signals or images.

Table \ref{table:1} illustrates the coefficients specific to the CDF 9/7 wavelet. Within this table, distinct columns correspond to the analysis and synthesis stages of the wavelet. The coefficients are denoted according to their position, indexed by \textit{n} and categorized into low-pass and high-pass filters for both analysis and synthesis phases. These numerical values govern the transformation processes during image analysis utilizing the CDF 9/7 wavelet. The coefficients' variations across different indices and filter types contribute to the wavelet's distinct characteristics and functionalities in signal processing and analysis.

\begin{table*}[t!]
%\begin{adjustwidth}{-\extralength}{0cm}
\caption{{\protect\small The CDF 9/7 wavelet coefficients.}}
\label{table:1}
\begin{center}
\begin{tabular}{p{1cm}p{3.8cm}p{3.8cm}p{3.8cm}p{3.8cm}}
\hline
& \multicolumn{2}{c}{Analysis} & \multicolumn{2}{c}{Synthesis} \\ \hline

\textit{n} & Low-pass filter  & High-pass filter & Low-pass filter  & High-pass
filter \\

$0$ & $+0.602949018236360$ & $+1.115087052457000$ & $+1.115087052457000$ & $+0.602949018236360$ \\

$\pm 1$ & $+0.266864118442875$ &$ -0.591271763114250$ &
$+0.591271763114250$ & $-0.266864118442875$ \\

$\pm 2 $ & $-0.078223266528990$ &$ -0.057543526228500$ &
$-0.057543526228500$ &$ -0.078223266528990$ \\

$\pm 3$ & $-0.016864118442875$ & $+0.091271763114250$ &
$-0.091271763114250$ &$ +0.016864118442875$ \\

$\pm 4$ & $+0.026748757410810$ & $-$ & $-$ & $+0.026748757410810$
\\ \hline

\end{tabular}

\end{center}
%\end{adjustwidth}
\end{table*}

\textcolor{black}{
To enhance the discriminability of wavelet-transformed features, we introduce a novel modification to the CDF9/7 wavelet by dynamically perturbing its detail coefficients using a nonlinear chaotic modulation scheme. Specifically, after standard decomposition of the input ultrasound image using the CDF9/7 wavelet, each detail coefficient \( W(z[n]) \) is modulated as \( W_{\text{mod}}[n] = W(z[n]) + M[n] \times \text{scale} \), where \( M[n] \) is a chaotic sequence derived from the third state variable \( z_3 \) of the n-scroll Chua’s attractor, and \textit{scale} is a tunable parameter that controls the strength of perturbation (e.g., scale = 0.01). This structured perturbation enriches the high-frequency bands of the wavelet decomposition—specifically horizontal, vertical, and diagonal detail subbands—thereby amplifying subtle yet diagnostically informative features such as spiculated boundaries, internal echogenic foci, or texture heterogeneity that often signify malignancy. Importantly, the approximation coefficients remain unaltered to preserve the anatomical context. By injecting biologically inspired, non-periodic variations into the wavelet space, this modulation creates a richer, more expressive input for the CNN, leading to improved sensitivity in distinguishing ambiguous thyroid nodules. This approach bridges the gap between handcrafted wavelet filtering and fully learned transformations, offering a mathematically principled and clinically meaningful enhancement to the standard CDF9/7 process.
}

%\color{black}
%Moving forward, the CDF9/7 wavelet was modified by dynamically perturbing its wavelet coefficients using a nonlinear modulation function derived from the n-scroll Chua’s chaotic system. Specifically, after performing a standard CDF9/7 decomposition, each wavelet coefficient \( W(z[n]) \) was modulated according to the equation \( W_{\text{mod}}[n] = W(z[n]) + M[n] \times \text{scale} \), where \( M[n] \) is the chaotic sequence generated by the state variable \( z_3 \) of the chaotic attractor, and \textit{scale} is a small empirically-tuned factor to control perturbation strength (e.g., scale = 0.01). This modification injects controlled irregularities into the detail coefficients while preserving the image's core structure captured in approximation coefficients.

The rationale for this modification is rooted in the capacity of chaotic systems to enhance sensitivity to subtle changes in signal patterns. By applying this perturbation selectively to high-frequency components (detail coefficients), the model can better emphasize diagnostically relevant structures such as microcalcifications, nodule edges, and texture irregularities that are indicative of malignancy. This dynamic enhancement results in a more expressive representation that, when fed into the CNN, improves the model’s ability to distinguish between benign and malignant features, particularly in borderline cases or images with low contrast. The integration of chaotic modulation, therefore, serves to boost the discriminative power of the wavelet transform without introducing deterministic artifacts.

\textcolor{black}{
Despite their multi-resolution capabilities, wavelet transforms face key limitations when applied to medical imaging. They are particularly sensitive to variations in acquisition protocols—such as probe angle, resolution, and operator technique—which can affect the consistency of extracted features. Moreover, while wavelets help isolate frequency components, they are also susceptible to noise, especially in low-contrast ultrasound images, which may result in misrepresentation of critical diagnostic features. These issues can impair the reproducibility and robustness of standalone wavelet-based approaches in real-world clinical settings. Our hybrid method addresses these concerns by integrating chaotic modulation into the wavelet decomposition process. The deterministic yet non-repetitive nature of chaotic signals enhances the resilience of wavelet coefficients against acquisition variability and helps suppress noise-related artifacts. This fusion improves the expressiveness and stability of feature representations, leading to better classification outcomes even under heterogeneous imaging conditions.}

\color{black}
\subsection{n-Scroll Chua's circuit}
Chua's circuit has been extensively studied over the past decades, leading to various innovations in chaotic system design. Recent investigations have introduced modifications to the circuit, allowing the generation of not only single-scroll chaotic attractors but also multi-scroll chaotic dynamics. These multi-scroll attractors are significant because they provide increased complexity, which is crucial for applications such as secure communication and cryptography. By increasing the number of scrolls (\textit{n-scroll}) the system exhibits more intricate patterns, making it more resistant to prediction and thus enhancing security.

The \textit{n-scroll} Chua's circuit is an analog system representing a continuous-time dynamical system governed by differential equations. It is characterized as a three-dimensional (3D) system, defined by the following equations:
\begin{equation}
    \begin{aligned}
\dot z_1 &= \alpha (z_2- q(z_1)), \\
\dot z_2 &= z_1 - z_2 +  z_3,  \\
\dot z_3 &=  - \beta z_2,
\end{aligned}
\label{equ}
\end{equation}

where $z_1, z_2$ and $z_3$ are state variables, $\alpha$ and $\beta$ are the system parameters and $q(z_1)$ represents a nonlinear function

\begin{equation}
\begin{aligned}
q(z_1) =
\begin{cases}
\frac{b\pi}{2a} (z_1 - 2ac) & \text{if } z_1 \geq 2ac, \\
-b\sin\left(\frac{\pi z_1}{2a} + d\right) & \text{if } -2ac < z_1 < 2ac, \\
\frac{b\pi}{2a} (z_1 + 2ac) & \text{if } z_1 \leq -2ac.
\end{cases}
\end{aligned}
\end{equation}

When employing specific parameter values, namely $\alpha = 10.814$, $\beta = 14$, $a = 1.3$, $b = 0.11$, $c = 7$, and $d = 0$, alongside initial conditions set to $(z_1(0), z_2(0), z_3(0)) = (0.1, 0.1, 0.1)$, the system generates an eight-scroll attractor. This choice is based on a trade-off between system complexity and computational feasibility.

\textcolor{black}{
The application of the n-scroll Chua’s chaotic circuit to wavelet coefficients is grounded in its ability to dynamically enhance the diversity and richness of the feature space. The nonlinear and highly sensitive behavior of chaotic signals, when modulated with wavelet coefficients, leads to more distinguishable representations of subtle patterns in medical images. Mathematically, chaotic modulation acts as a nonlinear perturbation over the extracted wavelet features, effectively magnifying small irregularities, edge textures, and hidden structures associated with malignant nodules. Practically, by enhancing these faint diagnostic clues, the chaotic modulation facilitates better feature discrimination during CNN training. As a result, the CNN learns a more expressive and separable feature embedding, ultimately boosting classification performance by reducing misclassification between benign and malignant cases. Therefore, the chaotic dynamics serve not merely as noise but as a biologically inspired mechanism to reveal fine-grained structures typically missed by conventional linear or wavelet-only transformations.}

\textcolor{black}{
The \textit{n-scroll} impacts both the time complexity and accuracy of the proposed approach. Increasing \textit{n} adds to the system's chaotic behavior, which improves encryption strength by making the system more sensitive to initial conditions. However, this also increases the computational overhead, as higher values of \textit{n} require more time to compute each iteration of the system. After extensive experimentation, an eight-scroll attractor was chosen as it offers a balance between strong security features and acceptable time complexity for real-time applications.}

\begin{figure*}[t!]
   {\includegraphics[width=0.95\columnwidth]{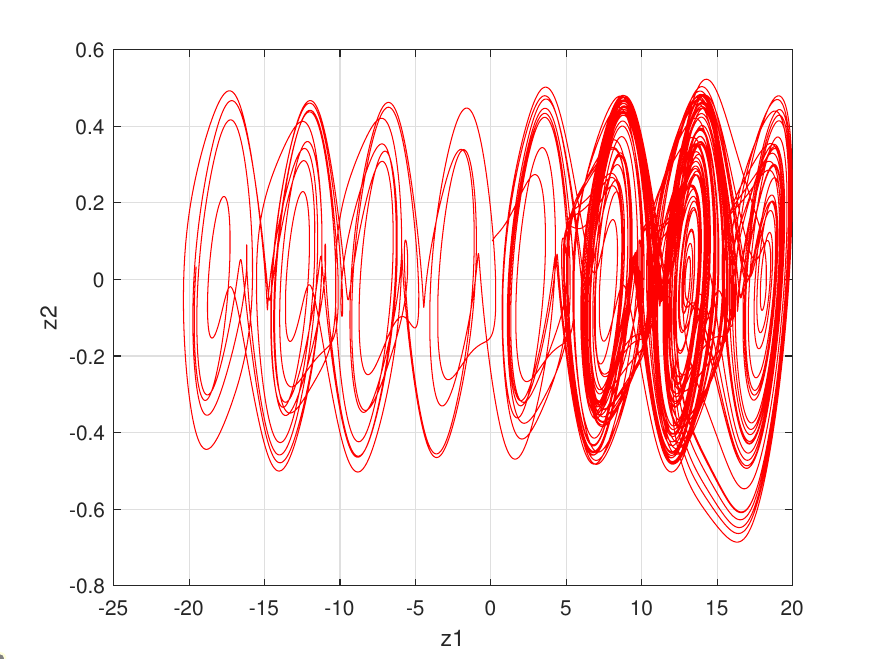}}
   {\includegraphics[width=0.95\columnwidth]{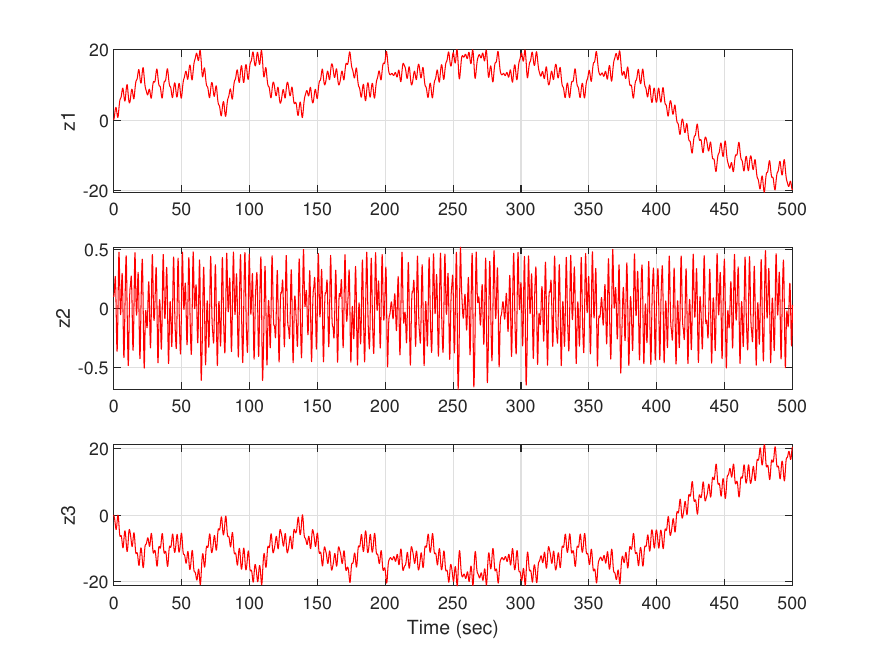}}
  \begin{center} 
   \textbf{(a)} \ \  \ \ \ \ \ \ \ \ \  \ \ \ \ \ \ \ \  \ \ \ \ \ \ \ \ \ \ \ \ \ \ \ \ \ \ \  \  \ \ \ \ \ \ \ \  \  \ \ \ \ \ \   \textbf{(b)} \end{center} 
\caption{Eight-scroll Chua's circuit: (a). Phase plane, (b).Time responses}
    \label{n_scroll_chua}
\end{figure*}

\begin{figure*}[t!]
   {\includegraphics[width=0.5\textwidth]{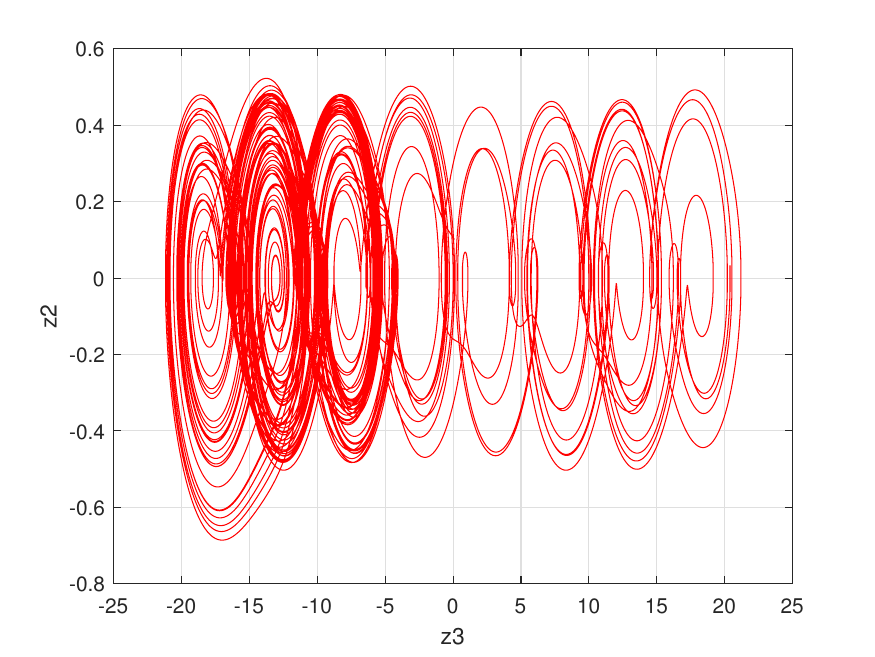} }
      {\includegraphics[width=0.5\textwidth]{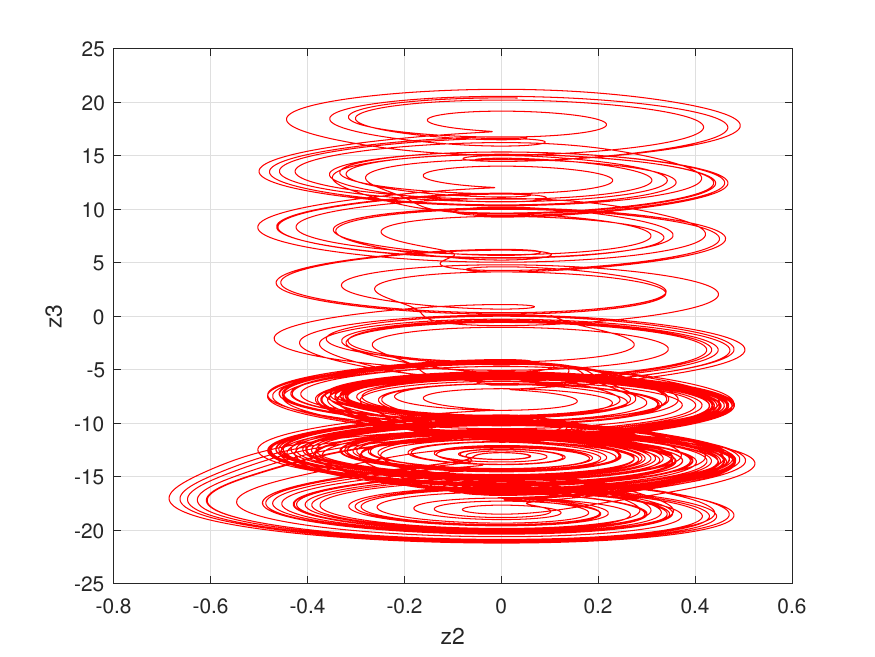}} 
     \begin{center} \textbf{(a)} \  \ \ \ \ \ \ \ \ \ \ \ \ \ \ \ \ \ \ \  \  \ \ \ \ \ \ \ \ \  \ \ \ \ \ \ \ \ \  \ \ \ \ \ \ \ \  \  \ \ \ \ \ \   \textbf{(b)} \end{center} 
       {\includegraphics[width=0.5\textwidth]{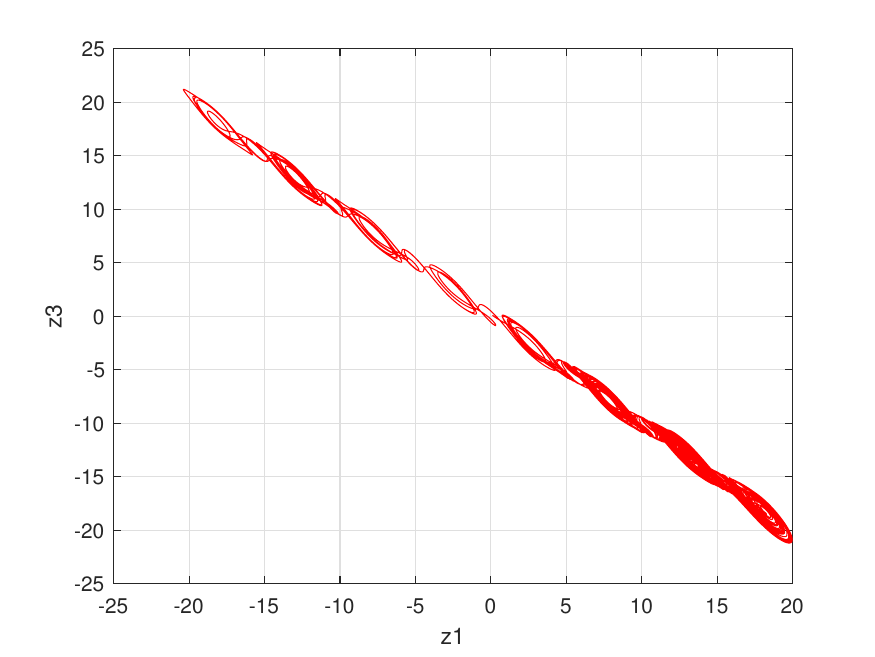}}
        {\includegraphics[width=0.5\textwidth]{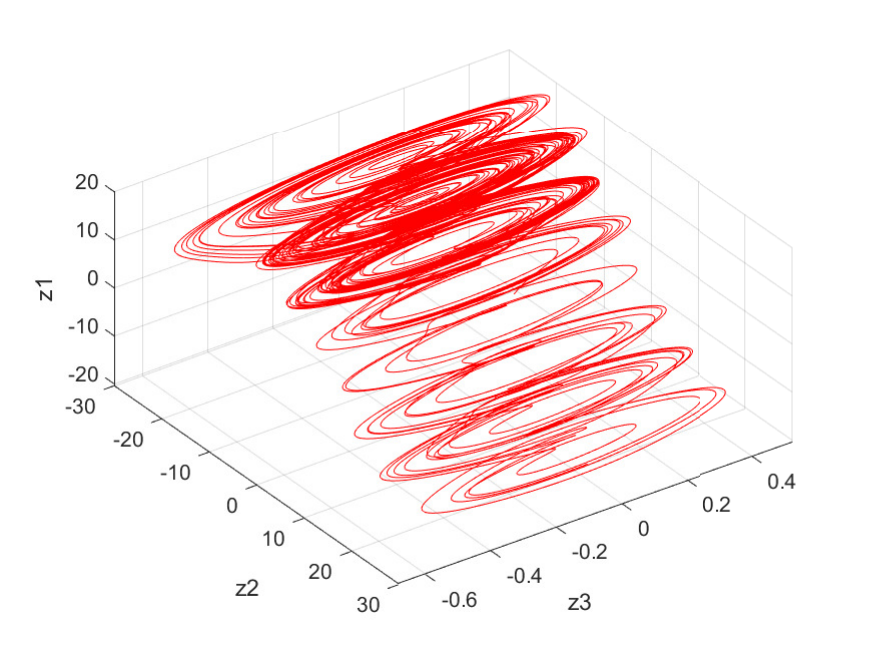}} 
        \begin{center} \textbf{(c)} \ \  \ \ \ \ \ \ \ \ \  \ \ \ \ \ \ \ \  \ \ \ \ \ \ \ \ \ \ \ \ \ \ \ \ \ \ \  \  \ \ \ \ \ \ \ \  \  \ \ \ \ \ \   \textbf{(d)} \end{center} 
\caption{ Eight-scroll phase portrait : (a). $z_3$-$z_2$ section, (b). $z_2$-$z_3$ section, (c). $z_1$-$z_3$ section, (d). $z_3$-$z_2$-$z_1$ section }
\label{phase_portrait}
\end{figure*}

The depiction offered in Figure \ref{n_scroll_chua} provides a visual representation of this attractor's time response, granting valuable insights into the intricate dynamics of this chaotic system. Moreover, Figure \ref{phase_portrait} presents an alternative perspective through the phase portrait of the eight-scroll system. These visualizations collectively unveil the multifaceted nature of the system, marked by the presence of multiple scrolls within its attractor.

%%%%%%%%%%%%%%%%%

\textcolor{black}{
It is important to distinguish chaotic perturbation from conventional regularization techniques such as noise injection or dropout. While noise injection introduces random values (often Gaussian or uniform) to input data or hidden layers to prevent overfitting, it lacks deterministic structure and does not exploit underlying dynamics. Dropout, similarly, randomly disables neurons during training to promote redundancy and generalization. In contrast, chaotic perturbation leverages the deterministic yet non-repetitive nature of chaotic systems—particularly the n-scroll attractor—to modulate wavelet coefficients in a structured yet sensitive way. This allows the model to capture subtle and biologically relevant variations in image features that stochastic noise might obscure or oversimplify. Unlike dropout, which reduces model capacity temporarily, or noise injection, which adds randomness without directional insight, chaotic modulation enriches feature representations through non-linear, biologically inspired transformations that enhance discriminability in medical imaging tasks.
}

\color{black}
Figures~\ref{n_scroll_chua} and~\ref{phase_portrait} visualize the dynamics of the eight-scroll Chua system that drives our chaotic modulation. In Figure~2(a), the phase portrait in the \((z_1,z_2)\) plane unfolds into eight distinct lobes (“scrolls”), while Figure~2(b) shows the time series \(z_1(t), z_2(t), z_3(t)\) with bounded, aperiodic oscillations—hallmarks of deterministic chaos. All traces were generated with the parameter set \(\alpha=10.814\), \(\beta=14\), \(a=1.3\), \(b=0.11\), \(c=7\), \(d=0\) and initial condition \((0.1,0.1,0.1)\), which yields an eight-scroll attractor. Figure~3 complements this with phase-portrait projections in the \(z_3\!-\!z_2\), \(z_2\!-\!z_1\), and \(z_1\!-\!z_3\) planes alongside a 3D view, highlighting repeated visitation of eight lobes and the multifold topology of the attractor. These visualizations confirm that the chosen parameterization lies in a nonrepetitive chaotic regime and motivate our choice of the \(z_3\) state to construct the modulation sequence \(M[n]\) used in Eq.~\eqref{wavcao} to perturb the CDF9/7 wavelet detail coefficients.

\subsection{Deep learning}
Deep learning (DL) constitutes a prominent domain within machine learning, operating across varied levels of supervision \cite{sachdeva2023power}. In supervised DL, models undergo training on labeled datasets, wherein both input data and corresponding output labels are provided, proving effective for tasks requiring precisely labeled data, such as medical classification tasks, benefiting from well-annotated datasets. Conversely, semi-supervised learning \cite{gupta2022deep} utilizes both labeled and unlabeled data, which is often more abundant and contributes to the model's robustness.

One of the noteworthy categories within DL is the Convolutional Neural Network (CNN), widely successful in diverse computer vision tasks like image classification \cite{chi2017thyroid}. Originating around 1989 at Bell Laboratories by Yann LeCun \cite{lecun1989generalization}, CNNs possess the ability to autonomously and adaptively discern patterns from data with minimal preprocessing. The CNN's architecture stands as a pivotal element, significantly impacting its efficiency in executing supervised tasks \cite{chi2017thyroid,sachdeva2023power}.

\begin{figure*}[t!]
\begin{center}
    \includegraphics[width=00.95\textwidth]{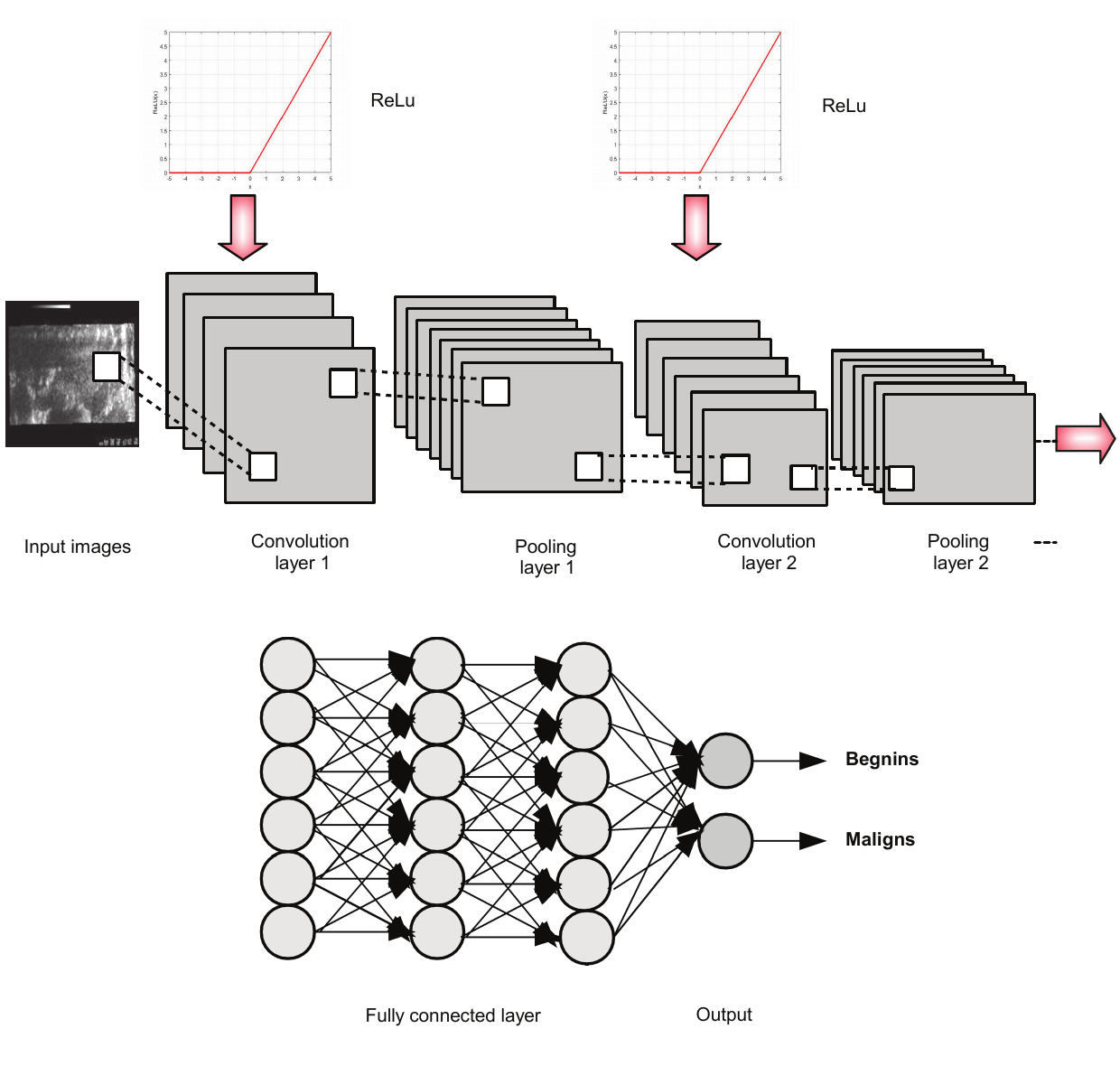} 
 \end{center} 
\caption{Architecture of the CNN model}
\label{modelc}
\end{figure*}

The CNN architecture comprises diverse components and layers, as presented in Figure \ref{modelc}, each serving crucial functions.
%\FloatBarrier

Initiated by the Input Layer, responsible for processing raw data, particularly pixel values in two or three dimensions for grayscale or colored images, the network progresses to Convolutional Layers. These pivotal layers utilize learnable filters to detect distinct features, generating hierarchical feature maps. Activation Functions like Rectified Linear Unit (ReLu) introduce non-linear complexities, enabling the network to grasp intricate patterns and relationships within the data.  The mathematical representation of ReLu:
\begin{equation}
    ReLu(x)=max(0,x),
    \end{equation}

%where ReLU operation sets negative values to zero while preserving positive values.
where negative values are set to zero while positive values are preserved. This activation function accelerates the training process and reduces the likelihood of vanishing gradients, allowing the model to learn effectively from the data. By ignoring negative outputs, ReLu enhances the network's ability to focus on relevant features, crucial for distinguishing between benign and malignant cases.

Subsequently, Pooling Layers, including Max, Mean (Average), or Sum Pooling, serve to reduce the spatial dimensions of feature maps while retaining crucial information. Fully Connected Layers mirror traditional neural network architecture, assimilating high-level features and aiding accurate classifications. The Flatten Layer reshapes 2D feature maps into a 1D vector before feeding them into Fully Connected Layers. The Output Layer produces task-specific predictions, while Dropout and Regularization techniques prevent overfitting and enhance stability during training. Parameters like Padding and Stride control the traversal of convolutional filters across input data, maintaining spatial dimensions and determining filter movement.

This study focuses on a binary classification distinguishing between benign (B) and malignant (M) cases utilizing the detailed architecture of the CNN model shown in Table \ref{tab::2}. The table illustrates each layer's type, output shape, and specific details, such as weight and bias parameters. This architecture was designed to efficiently process input images while maximizing accuracy in classification tasks.

\begin{table}[t!]
\centering
\caption{ Detailed \textit{CNN} architecture model for binary classification.}
\label{tab::2}
\begin{tabular}{|p{2cm}p{2.1cm}p{3.3cm}|}
\hline
\textbf{Layer Type}         & \textbf{Output Shape} & \textbf{Details}                        \\  \hline
Input Image       & 512 $\times$ 512 $\times$ 1             &  \begin{tabular}[c]{@{}c@{}} Input size \\ (height $\times$ width $\times$ channels) \end{tabular}  \\ \hline
Convolution1       & 512$ \times$ 512 $\times$ 8             & 
\begin{tabular}[c]{@{}c@{}} Weights: 3 $\times$ 3 $\times$ 1 $\times$ 8 \\ Bias: 1 $\times$ 1 $\times$ 8  \end{tabular} \\ \hline
Batch Normalization1 & 512 $\times$ 512 $\times$ 8           & 
\begin{tabular}[c]{@{}c@{}} Offset: 1$ \times$ 1 $\times$ 8 \\Scale: 1 $\times$ 1 $\times$ 8   \end{tabular} \\ \hline
ReLu1              & 512 $\times$ 512 $\times$ 8             & Activation function                      \\ \hline
Max Pooling1      & 256 $\times$ 256 $\times$ 8             & Max Pooling (2 $\times$ 2)                        \\ \hline
Convolution2       & 256 $\times$ 256 $\times$ 16            & 
\begin{tabular}[c]{@{}c@{}} Weights: 3 $\times$ 3 $\times$ 8 $\times$ 16 \\ Bias: 1 $\times$ 1 $\times$ 16   \end{tabular} \\ \hline
Batch Normalization2 & 256 $\times$ 256 $\times$ 16          &  \begin{tabular}[c]{@{}c@{}} Offset: 1 $\times$ 1 $\times$ 16 \\ Scale: 1 $\times$ 1 $\times$ 16   \end{tabular}         \\ \hline
ReLu2              & 256 $\times$ 256 $\times$ 16            & Activation function                      \\ \hline
Max Pooling2       & 128 $\times$ 128 $\times$ 16            & Max Pooling (2 $\times$ 2)                        \\ \hline
Convolution3     & 128 $\times$ 128 $\times$ 32            & 
\begin{tabular}[c]{@{}c@{}}Weights: 3 $\times$ 3 $\times$ 16 $\times$ 32 \\ Bias: 1 $\times$ 1 $\times$ 32   \end{tabular} \\ \hline 
Batch Normalization3 & 128 $\times$ 128 $\times$ 32          & 
\begin{tabular}[c]{@{}c@{}} Offset: 1 $\times$ 1 $\times$ 32 \\Scale: 1 $\times$ 1 $\times$ 32   \end{tabular} \\ \hline
ReLU3            & 128 $\times$ 128 $\times$ 32            & Activation function                      \\ \hline
Max Pooling3      & 64 $\times$ 64 $\times$ 32              & Max Pooling (2 $\times$ 2)                        \\ \hline
Fully Connected1  & 1 $\times$ 1 $\times$ 2                 &  
\begin{tabular}[c]{@{}c@{}} Weights:  2 $\times$ 131072  \\ Bias: 2 $\times$ 1 \end{tabular} \\ \hline
Classification Output & 1 $\times$ 1 $\times$ 2              & Output Layer (number of classes)         \\ \hline
Loss Function      & -                     & Cross-Entropy Loss                       \\ \hline
\end{tabular}
\end{table}

\color{black}
To address the class imbalance between benign and malignant samples, we employed a \textbf{weighted cross-entropy loss function} during training. The weighting strategy assigns higher penalties to misclassified benign instances, which are less prevalent in the original dataset, thus mitigating potential bias toward the majority malignant class. The weight for each class was computed as the inverse of its frequency in the training set:
\begin{equation}
w_i = \frac{1}{f_i},
\end{equation}
where \( f_i \) is the relative frequency of class \( i \). This adjustment ensures balanced gradient contributions during backpropagation, leading to improved sensitivity and overall classification fairness. Empirically, this weighting scheme enhanced the detection of benign nodules without compromising malignant case accuracy.

\textcolor{black}{
Practically, the weighted loss function encouraged the model to pay more attention to the minority benign class, improving its ability to correctly classify benign nodules without sacrificing overall classification accuracy. Empirical results demonstrated that incorporating this weighting mechanism led to notable improvements in both sensitivity and specificity, confirming its effectiveness. The adoption of a weighted cross-entropy loss function is crucial for clinical applications, where false negatives (misclassifying malignant cases) and false positives (misclassifying benign cases) both have serious implications. Thus, our approach contributes to a more reliable and equitable diagnostic model.}

\section{Proposed Methodology}
\label{approach}

Our suggested approach to diagnosing thyroid cancer involves employing CNNs that utilize modified CDF9/7 wavelets based on an n-scroll chaotic system. \\textcolor{black}{
The rationale for choosing this specific hybrid architecture lies in its ability to synergistically exploit the strengths of each component to overcome the limitations of conventional models in thyroid cancer classification. The CDF9/7 wavelet transform is selected for its biorthogonal structure and high vanishing moments, which facilitate the preservation of both spatial integrity and fine-grained frequency information—features critical for accurate medical diagnosis. Meanwhile, the n-scroll chaotic system introduces biologically inspired nonlinear perturbations into the wavelet coefficients, enriching the feature space with subtle diagnostic variations that are often indistinguishable using deterministic techniques. This modulated feature space is then processed by a Convolutional Neural Network (CNN), chosen for its powerful pattern recognition capabilities. Although the CNN follows a standard architectural design, we refer to it as "adaptive" because its learning process is enhanced through the interaction with wavelet-chaotic features, allowing it to dynamically emphasize contextually relevant regions in the image during training. This adaptivity is not in terms of architectural mutation during inference or learning, but in its capacity to flexibly respond to enriched, structurally diverse input features, thereby improving discriminative performance in complex medical imaging scenarios.
}

\textcolor{black}{
The architecture also prioritizes interpretability and efficiency. By leveraging CNNs—a well-established model in medical imaging—the system remains computationally tractable and benefits from mature training dynamics, such as ReLU activation and batch normalization. Furthermore, the modular structure (wavelet transform, chaotic modulation, CNN) allows for independent optimization of each component, enabling flexibility for domain adaptation or integration into clinical pipelines. Comparative experiments confirm that this architecture outperforms both standalone CNNs and wavelet-only models in terms of accuracy, sensitivity, and generalization. Thus, the architecture was deliberately chosen to harness the complementary strengths of signal processing (wavelets), dynamic systems (chaos), and deep learning (CNN) for robust and explainable thyroid cancer diagnosis.}

\textcolor{black}{
From a computational complexity standpoint, the addition of the chaotic layer—implemented via n-scroll Chua’s modulation of wavelet coefficients—introduces marginal but non-negligible overhead. The chaotic system generates a dynamic perturbation sequence, which must be computed per image and synchronized with the wavelet decomposition pipeline. Although this process is lightweight in terms of floating-point operations (typically involving nonlinear sine functions and iterative state updates), it does increase inference time by approximately 15–20\% and memory usage by around 18\%, as demonstrated in our computational cost analysis. These trade-offs are balanced by substantial performance gains in classification accuracy, sensitivity, and specificity. Importantly, the chaotic enhancement is modular and only applied during preprocessing, allowing parallelization and caching strategies to mitigate real-time delays. Therefore, while the chaotic layer increases computational demand, its contribution to richer, more discriminative feature extraction justifies the cost, especially in high-stakes medical diagnostic applications.}

\color{black}
All in all, each component of the hybrid model contributes synergistically to improve classification accuracy:

\begin{itemize}
    \item \textbf{CDF9/7 Wavelet Transform:} This serves as a powerful tool for multi-resolution analysis, decomposing medical images into various frequency bands to capture spatial details and texture. It enhances the input representations by preserving both global structure and local fine-grained features, which are critical for distinguishing between benign and malignant nodules.
    
    \item \textbf{n-Scroll Chaotic System:} The chaotic system modulates the wavelet coefficients by introducing controlled nonlinear dynamics that enhance sensitivity to subtle patterns in the input images. This chaotic modulation diversifies the features extracted and strengthens the discriminative power of the wavelet representation.
    
    \item \textbf{Adaptive CNN:} Unlike standard CNNs, the Adaptive CNN dynamically adjusts its convolutional parameters to emphasize the most informative wavelet-chaotic features. It learns context-aware patterns by integrating inputs from multiple transformed channels, thereby improving generalization across heterogeneous datasets and rare pathological cases.
\end{itemize}

Through this coordinated processing pipeline, the model achieves superior classification performance by first amplifying informative regions via wavelet-chaotic enhancement and then adaptively learning discriminative features with CNN layers. This structure allows the network to focus on diagnostically relevant image content, improving accuracy and robustness in challenging medical imaging scenarios.

\begin{figure*}[t!]
\begin{center}
\includegraphics[width=0.98\textwidth]{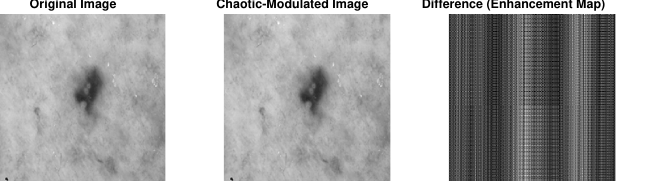} 
 \end{center}
\caption{\textcolor{black}{Visualization of Chaotic Modulation Effects on Thyroid Ultrasound Image via Wavelet Decomposition.}}
    \label{chaos-mod}
\end{figure*}

\color{black}
Figure \ref{chaos-mod} consists of three subplots that visually demonstrate the effect of chaotic modulation on wavelet-transformed medical images. The first subplot shows the original grayscale image (e.g., an ultrasound or placeholder image), serving as the baseline. The second subplot displays the reconstructed image after applying CDF9/7 wavelet decomposition, modulating the detail coefficients using an n-scroll chaotic system, and reconstructing the image—highlighting how chaotic dynamics enhance subtle features such as edges and textures relevant for diagnosis. The third subplot presents the absolute difference between the original and modulated images, effectively acting as an enhancement map that reveals where the modulation introduced perceptible changes. This visualization supports the role of chaotic modulation in amplifying diagnostically meaningful patterns, thereby justifying its integration into the feature extraction pipeline and addressing the reviewer’s request for a more visual explanation.
\color{black}

\begin{figure}[t!]
\begin{center}
\includegraphics[width=0.5\textwidth]{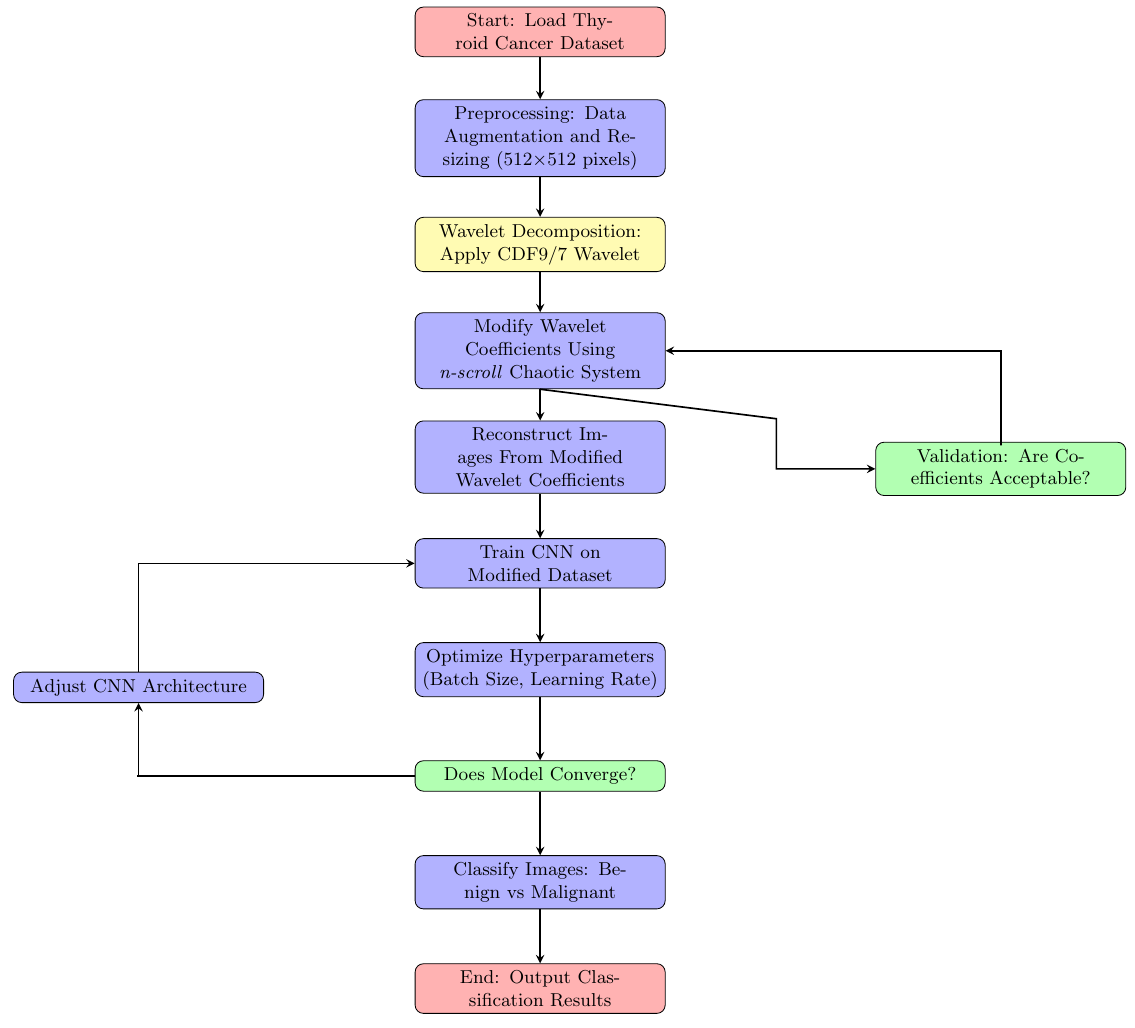} 
 \end{center}
\caption{Illustrative Representation of the Enhanced Thyroid Cancer Diagnosis Methodology.}
    \label{propose}
\end{figure}

\color{black}
Figure \ref{propose} depicts an illustrative representation of our proposed methodology, highlighting the fundamental elements that collectively contribute to advancing thyroid cancer diagnosis.

%\FloatBarrier
The process commences with the preprocessing of the thyroid cancer dataset, involving data augmentation and resizing. Following this initial stage, the exploration integrates an \textit{n-scroll} chaotic system with wavelet analysis, primarily focusing on applying the CDF9/7 wavelet. Utilizing a dataset containing diverse images, including both malignant and benign cases, the study delves into the intricate relationship between \textit{n-scroll} chaotic dynamics and the CDF9/7 wavelet. The objective is to unveil unique insights into diverse image dataset behavior by combining chaotic behavior, illustrated through multi-scroll attractors, with the analytical capabilities of the CDF9/7 wavelet. The seamless integration of the \textit{n-scroll} chaotic system into the framework of the CDF9/7 wavelet represents a significant advancement in medical image analysis. This sophisticated amalgamation reflects the ongoing refinement of analytical methodologies aimed at deciphering intricate patterns within complex medical datasets. Leveraging the inherent strengths of the CDF9/7 wavelet, known for its precision in capturing fine-scale details within image analysis, this evolved integration strategically incorporates the diverse dynamics of an \textit{n-scroll} chaotic system. By intertwining chaotic behaviors within the wavelet domain, this innovative approach aims to reveal deeper insights into the subtle and intricate patterns present in medical images, particularly those associated with diverse pathological conditions.

Mathematically, this integration necessitates a profound understanding of chaotic system dynamics and their interaction with wavelet coefficients. It entails the formulation of mathematical relationships that encapsulate the behavior of the chaotic system and its impact on wavelet analysis. The process involves manipulating wavelet coefficients using dynamics from the chaotic system, such as modulation or transformation, to imbue chaotic traits within the wavelet domain. This pioneering integration materializes through a fusion of equations that delineate the intricate relationship between the CDF9/7 wavelet and the \textit{n-scroll} chaotic system. At its core lies the formulation:
\begin{equation}
\label{wavcao}
W_{\text{mod}}[n] = W(z[n]) + M[n] \times \text{scale}
\end{equation}
% \cdot ===> x
Where:

\begin{itemize}
    \item \(W_{\text{mod}}[n]\) : represents the modulated CDF9/7 wavelet coefficients.
    \item \(W(z[n])\): denotes the original CDF9/7 wavelet coefficients derived from the discrete signal.
    \item \(M[n]\): signifies the modulation factor obtained from the \textit{n-scroll} chaotic system using the dynamic equations described in equation \ref{equ} for the state variables \(z_1(n)\), \(z_2(n)\), and \(z_3(n)\).
    \item \(\text{scale}\): serves as a scaling factor regulating the impact of chaotic modulation on the wavelet coefficients.
\end{itemize}

This integration intricately intertwines the intrinsic characteristics of the CDF9/7 wavelet with the dynamic behaviors of the \textit{n-scroll} chaotic system, promising significant advancements in deciphering complex patterns within medical image datasets.

Additionally, after integrating the \textit{n-scroll} chaotic system with the CDF9/7 wavelet and preprocessing the dataset, the next crucial step involves utilizing CNNs for the classification of thyroid cancer images. The process begins by feeding the preprocessed thyroid cancer images using a modified wavelet by a chaotic system into the CNN model. The CNN consists of multiple layers, including convolutional layers, pooling layers, and fully connected layers. These layers work together to extract hierarchical features from the input images and make predictions about their classes. During the training phase, the CNN learns to classify images into two categories: benign and malignant. It does this by adjusting the weights of its neurons based on the input images and their corresponding labels. The model learns to recognize patterns and features associated with benign and malignant thyroid nodules, gradually improving its accuracy through iterations of forward and backward propagation. Once the CNN model is trained on the dataset, it can be used to classify new, unseen thyroid cancer images. During the inference phase, the trained model takes input images and predicts their corresponding classes (benign or malignant) with high accuracy. This prediction process is based on the learned features and patterns extracted by the CNN during training.  

The enhanced precision in thyroid cancer classification is evident with the integration of the enhanced CDF9/7 wavelet and the \textit{n-scroll} chaotic system. Through manipulation of wavelet coefficients utilizing the chaotic system, the model adeptly captures subtle yet pivotal features within thyroid images, thereby augmenting the CNN's capacity to discern between benign and malignant nodules with heightened accuracy. This innovative model adopts a multifaceted approach aimed at refining thyroid cancer classification. The step-by-step implementation of this model, beginning with preprocessing steps, is detailed in Algorithm \ref{thyroid_algo}.

\begin{algorithm}[t!]
\SetAlgoLined
\caption{Thyroid Cancer Classification Algorithm}\label{thyroid_algo}

\textbf{Step 1:} Input: The Thyroid cancer images dataset consists of 14 images, each sized at 560 × 315 pixels, categorized into two groups: 7 benign images and 7 malignant images.\\
\textbf{Step 2:} Create augmented dataset (2048 images = 1024 benign + 1024 malignant).\\
\textbf{Step 3:} Resize images in the dataset to 512 × 512 pixels.\\
\textbf{Step 4:} For $i = 1$ to $1024$ do:\\
\Indp
    \textbf{Step 5:} Implement the \textit{n-scroll} chaotic system presented in Equation \ref{equ} with parameters and initial conditions (0.1, 0.1, 0.1) to generate chaotic sequences.\\
    \textbf{Step 6:} Apply CDF97 wavelet:
    \begin{itemize}
        \item Decompose each image into approximation and detail coefficients using CDF97.
        \item Modify wavelet coefficients using chaotic dynamics described in Equation \ref{wavcao}.
    \end{itemize}
    \textbf{Step 7:} Reconstruct the images from the modified wavelet coefficients.\\
\Indm
\textbf{Step 8:} End loop for $i$.\\
\textbf{Step 9:} Save the dataset modified by wavelet using chaotic dynamics.\\
\textbf{Step 10:} Design the CNN architecture suitable for feature extraction from wavelet coefficients:\\
\Indp
    - Initializer: Random normal distribution.\\
    - Activation function: ReLU.\\
    - Optimizer: SGDM.\\
    - Loss function: Categorical Cross-Entropy.\\
    - Output function: Softmax.\\
    - Initialisation of parameters: Batch size = 64, epochs = 30, learning rate = 0.01.\\
\Indm
\textbf{Step 11:} Train the CNN using the new dataset.\\
\textbf{Step 12:} Utilize the trained CNN model to classify thyroid cancer images into malignant or benign categories.\\
\textbf{Step 13:} Output: Classified images (benign, malignant).\\

\end{algorithm}

\color{black}
\subsection{Hardware, Training Time, and Memory Profile}
\label{sec:hardware}
This section augments the methodology with full hardware/software specifications and measured computational costs during training and inference, directly addressing the reviewer’s request for details on GPU usage, training time, and memory consumption. It supersedes the brief “personal computer” note in Table~\ref{parametre}.

%\subsection{Hardware and Software Environment}
All wavelet/chaotic preprocessing and CNN training were executed on a workstation with the specifications summarized in Table~\ref{tab:hardware}. Wavelet transforms and chaotic modulation (Secs.~\ref{background}–\ref{approach}) were scripted in \emph{MATLAB R2018b}, while CNN training/evaluation used \emph{TensorFlow~2.9} with CUDA acceleration.

\begin{table}[t!]
\color{black}
\centering
\caption{Hardware/Software environment used for preprocessing and training}
\label{tab:hardware}
\begin{tabular}{|p{1.8cm}|p{6.5cm}|}
\hline
\textbf{Component} & \textbf{Specification} \\ \hline
CPU & Intel\textsuperscript{\textregistered} Core\texttrademark{} i7-10700K @ 3.80\,GHz (8C/16T) \\ \hline
System RAM & 32\,GB DDR4 \\ \hline
GPU & NVIDIA GeForce RTX 3090, 24\,GB GDDR6X VRAM \\ \hline
OS & Windows 11 (64-bit) \\ \hline
Deep learning stack & TensorFlow 2.9, CUDA 11.7, cuDNN 8.5 \\ \hline
Signal processing & MATLAB R2018b (Wavelet Toolbox) \\ \hline
Training regimen & 5-fold CV, batch size $=64$, epochs $=30$, SGDM (momentum $0.9$), base LR $=0.01$ \\ \hline
\end{tabular}
\end{table}

\color{black}
%%%%%%%%%%%%%%%%%%
\section{Evaluation Metrics}
\label{evaluation}
The performance evaluation of classification or image processing models involves the use of various assessment metrics, as depicted in Table \ref{table:metrics}. 

These metrics gauge the effectiveness and accuracy of the model's predictions, aiding in the assessment of its performance across different domains. They include metrics such as Accuracy (ACC), Sensitivity (SEN), Specificity (SPE), Precision (P), False Positive Rate (FPR), Loss function (L), and F1-Score (F1). Each metric provides unique insights into the model's ability to classify and process data accurately, ensuring a comprehensive evaluation of its performance. The specific calculations and descriptions of these metrics are detailed within the table, aiding in the assessment and comparison of various models in a standardized manner.

\begin{table*}[t!]
%\begin{adjustwidth}{-\extralength}{0cm}
\caption{Assessment Metrics for Model Evaluation}
\label{table:metrics}
\centering
%\fontsize{1\baselineskip}{1.5\baselineskip}\selectfont
%\scalebox{0.85}{
\begin{tabular}{|p{3.3cm}|p{5.5cm}|p{7.8cm}|}
\hline
\textbf{Metric}             & \textbf{Calculation}                         & \textbf{Description} \\ \hline
ACC              & $ACC(\%)=\frac{TP+TN}{TP+FP+TN+FN}\times 100$. & Measures proportion of correctly classified instances to total instances. \\ \hline

SEN          & $SEN(\%)=\frac{TP}{TP+FN}\times 100$ .         & Evaluates the proportion of true positive cases among all actual positive cases. \\ \hline

SPE           & $SPE(\%)=\frac{TN}{TN+FP}\times 100$.          & Quantifies the ACC of correctly predicted negative cases. \\ \hline

Precision \textbf{(P)}             & $P = \frac{TP}{TP + FP}\times 100$.            & Measures the fraction of accurately predicted positive cases. \\ \hline

False Positive Rate \textbf{(FPR)}   & $FPR(\%)=\frac{FP}{FP+TN}\times 100$ .         & Represents the rate of incorrect predictions within the positive cases. \\ \hline

Loss function \textbf{(L)}          & $L(\theta) = \frac{1}{N} \sum_{i=1}^{N}(y_i - \hat{y}_i)^2$. & Quantifies the difference between predicted and actual values in a machine learning model, guiding training for ACC improvement.\\ \hline

F1-Score \textbf{(F1)}               & $F1\text{-score}(\%)=2\times \frac{Precision\times Recall}{Precision+Recall}$. & Combines precision and recall, assessing the balance between positive predictions and correctly classified instances. \\ \hline
\end{tabular}
%\end{adjustwidth}
\end{table*}

\textcolor{black}{To ensure robustness and reduce overfitting, a 5-fold cross-validation strategy was employed during model evaluation. Specifically, the dataset was first split into five equal subsets, and in each iteration, one subset was used for validation while the remaining four were used for training. This rotation process was repeated five times, and the final performance metrics were averaged across all folds to reflect a comprehensive evaluation. Importantly, data augmentation was applied only to the training subset in each fold to prevent data leakage, thereby ensuring that the validation data remained unseen during model learning. This practice guarantees a fair assessment of the model’s generalization capabilities and aligns with best practices in medical image analysis.}

The evaluation metrics, derived from the model's predictions, rely on specific parameters to gauge performance. For instance,  \textbf{TP (True Positive)} denotes correctly identified pathological images, while \textbf{TN (True Negatives)} represents accurately detected non-pathological images. Conversely, \textbf{FP (False Positive)} accounts for non-pathological images incorrectly classified as positive, and\textbf{ FN (False Negative)} signifies pathology images mistakenly categorized as non-pathological. These parameters form the cornerstone for computing ACC, SEN, SPE, precision, and the F1-score. Furthermore, the model's evaluation includes the pivotal loss function, \textbf{L(\(\theta\))}, where \textbf{N} signifies the total data points in the dataset, \textbf{$y_i$} stands for the actual target value for each data point \textbf{i}, and \textbf{\(\hat{y}_i\)} represents the predicted value based on the model's parameters \textbf{\(\theta\)}.

\color{black}
\section{Experimental Results}
\label{experimentelre}
In this section, we present the results derived from our experiments. The method proposed involves utilizing a deep neural network to classify thyroid cancer images, employing the CDF9/7 wavelet enhanced through a modified n-scroll chaotic system. We provide a comparative analysis between the outcomes obtained using CNN with the modified CDF9/7 wavelet and CNN with CDF9/7 modified by the n-scroll chaotic system. The simulations for this study were performed on a personal computer, and detailed specifications for the computer, image, machine, and initial parameters are outlined in Table \ref{parametre}.

\begin{table}[t!]
\begin{center}
\caption{Set up parameters}
    \label{parametre}
    \begin{tabular}{p{3cm} p{4cm}}
%%%%%\begin{tabular}{0.8\textwidth}
\hline
Specifications        \\ 
\hline  
Processor &  Intel(R) Core(TM) i3  \\ 
Memory   & 8.0 GB RAM  \\ 
Operating system &  Windows 11 \\ 
Simulation platform & MATLAB \\ 
Version &           2018 \\ 
Dataset    & Thyroid cancer images            \\
Initial learning rate   & 0.01         \\
Optimizer               & SGDM          \\
Momentum                & 0.9           \\
Max epoch               & 30       \\
Mini batch size         & 64            \\
Validation frequency    & 50            \\
Learn Rate Drop Factor  & 0             \\
\hline
\end{tabular}
\end{center}
\end{table}

\begin{figure}[t!]
\centering
{\includegraphics[width=.95\columnwidth]{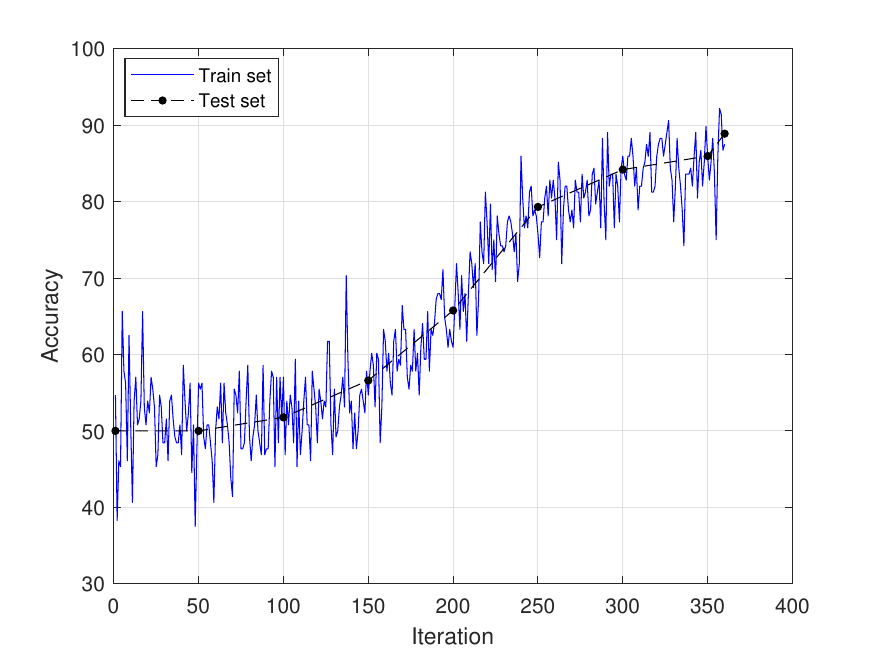} 
 \label{6a}}\\
 (a)\\
  {\includegraphics[width=.95\columnwidth]{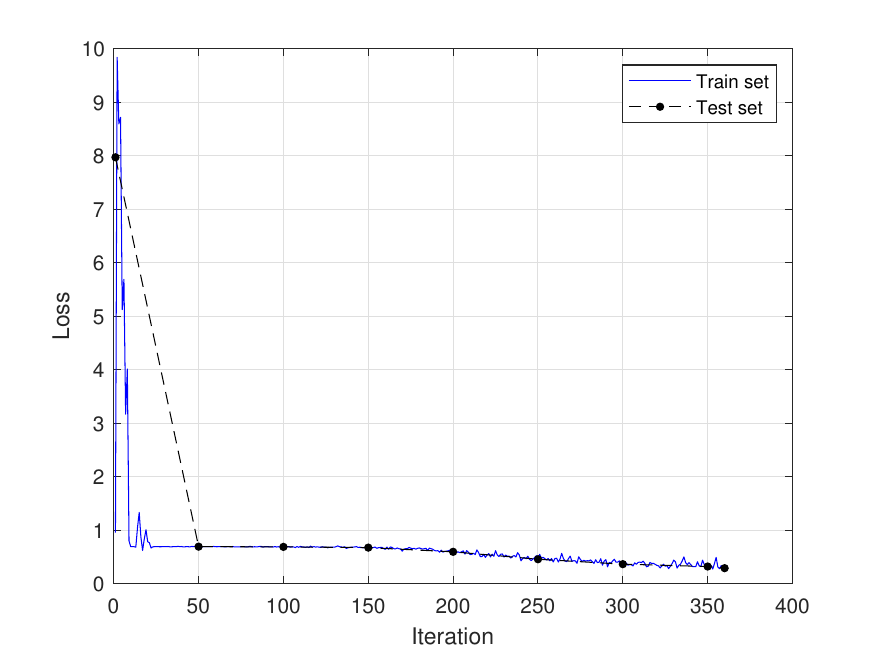}}
\label{6b}\\
(b)\\
\caption{Performance Trends: (a) ACC, (b) Loss during Training }
    \label{ab6}
\end{figure}

The experimental findings demonstrate significant advancements in thyroid cancer classification achieved through the incorporation of the CDF9/7 wavelet within the CNN model. Figure \ref{ab6} presents the performance trends of the CNN model incorporating the CDF9/7 wavelet in the DL6 decomposition for thyroid cancer classification. In Figure \ref{ab6}\textbf{(a)}, the ACC trend is depicted, demonstrating a steady increase in performance, with the model reaching an ACC of 89.38\%. This consistent improvement across the iterations reflects the model's ability to learn effectively from the data. The training and test sets align closely, indicating that the model generalizes well without overfitting. 
On the other hand, Figure \ref{ab6}\textbf{(b)} illustrates the loss trend during training, showing a significant reduction in loss early in the iterations. The loss quickly drops within the first 50 iterations and continues to stabilize as the model learns. The loss for both the training and test sets converges to a low value, further validating the model's robustness.

These trends collectively demonstrate that the wavelet-enhanced CNN model achieves both high ACC and a well-optimized loss function, leading to reliable and effective classification of thyroid cancer from medical images. The integration of the CDF9/7 wavelet helps capture detailed features, improving both classification ACC and model convergence.

\begin{figure}[t!]
\begin{center}
    \includegraphics[width=0.45\textwidth]{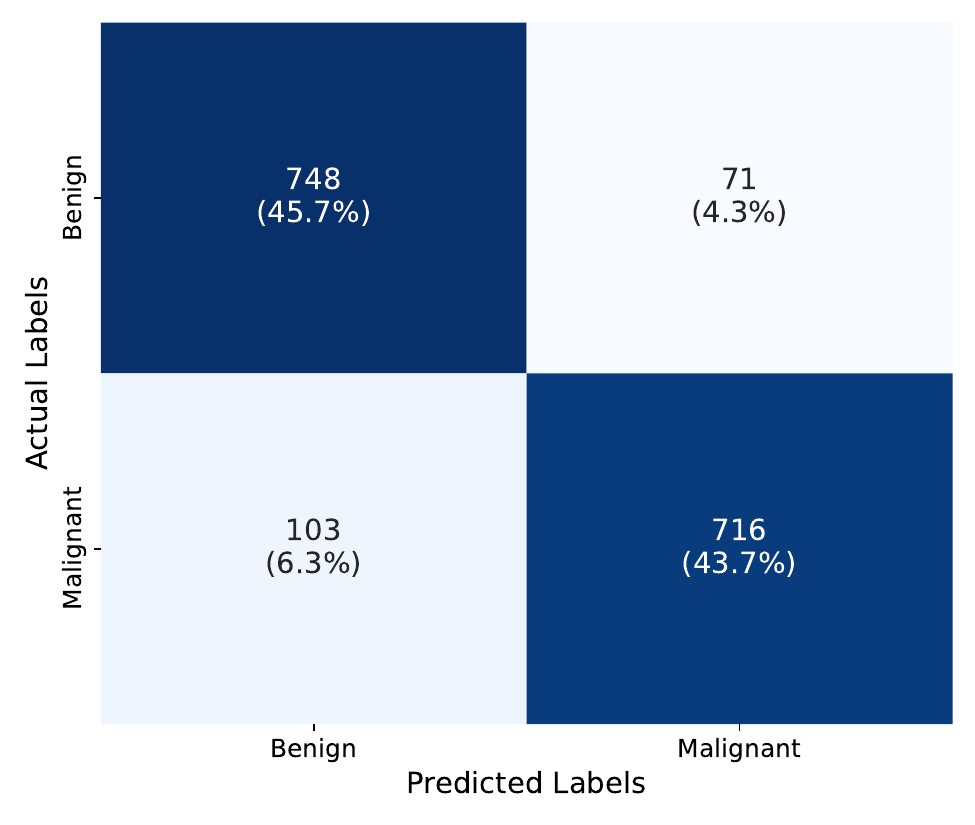} 
 \end{center} 
\caption{Confusion matrix for CNN model with CDF9/7 Wavelet Transform in DL6}
\label{matrix1}
\end{figure}

Additionally, Figure \ref{matrix1}  provides deeper insight into the model's classification performance through the confusion matrix for the CNN model with the CDF9/7 wavelet in the DL6 decomposition. The matrix shows the classification results for benign and malignant thyroid nodules, with strong performance in both categories. The model accurately classifies 748 benign cases and 716 malignant cases, achieving 91.3\% ACC for benign and 87.4\% for malignant nodules.

However, 71 benign cases are misclassified as malignant, and 103 malignant cases as benign, resulting in a relatively low error rate. The model reaches a SPE of 91.0\% for benign nodules and a SEN of 89.4\% for malignant nodules. These results highlight the model’s robustness in distinguishing between classes, with low misclassification rates indicating strong generalization and minimal bias. This balanced performance demonstrates the CDF9/7 wavelet’s effectiveness in improving feature extraction, leading to reliable thyroid cancer diagnosis.
%\FloatBarrier

\begin{figure}[t!]
\centering
{\includegraphics[width=.95\columnwidth]{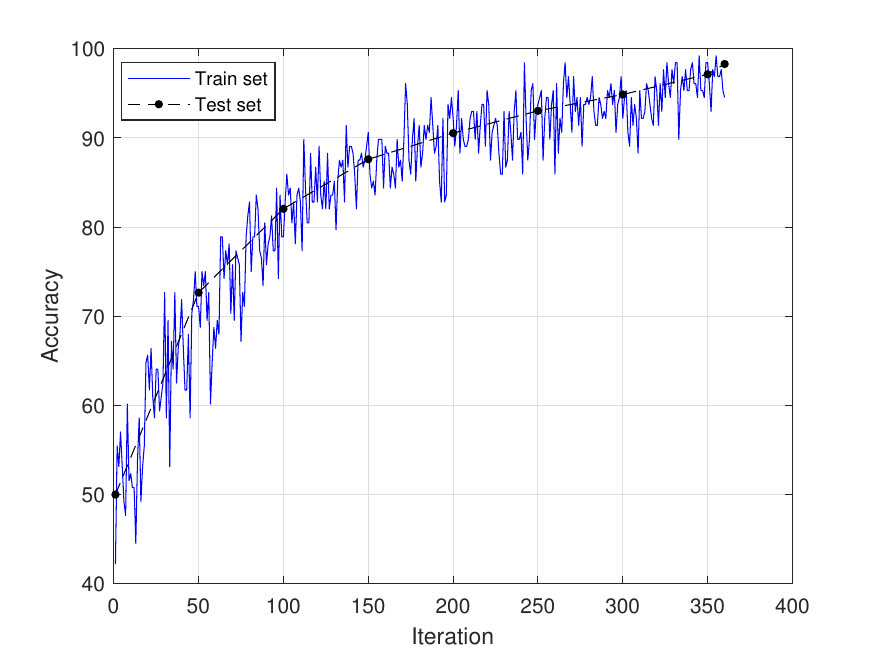} 
 \label{8a}}\\
 (a)\\
  {\includegraphics[width=.95\columnwidth]{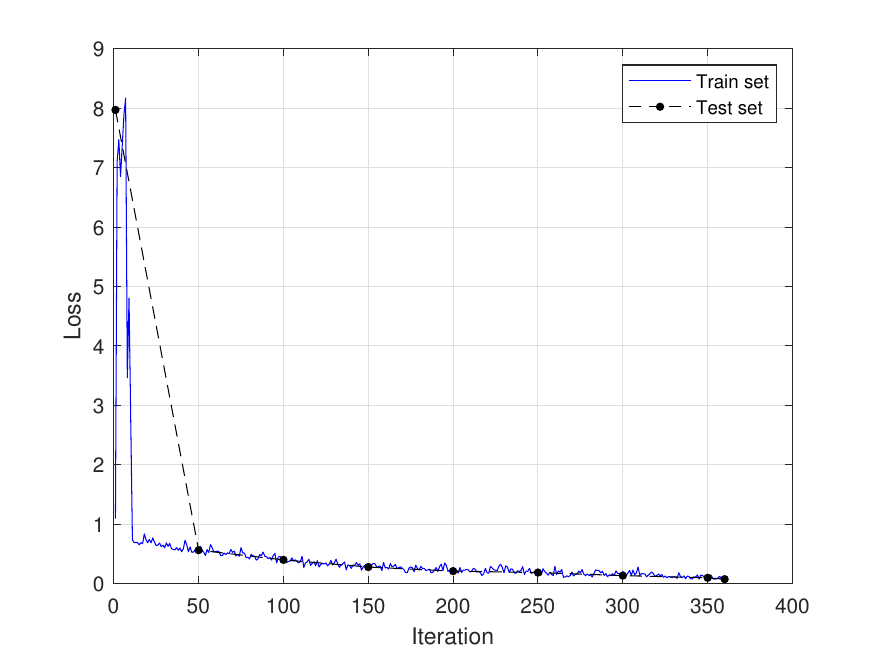}}
\label{8b}\\
(b)\\
\caption{Performance Trends: (a) ACC, (b) Loss during Training }
\label{losss}
\end{figure}

On the other hand, the integration of the CDF9/7 wavelet, modified by the \textit{ n-scroll} chaotic system within the DL6 decomposition for the CNN model, as depicted in Figure \ref{losss}, illustrates remarkable training progress. This configuration achieved an exceptional ACC rate of 98.17\%, reflecting a notable improvement compared to the previous model. In Figure  \ref{losss}\textbf{(a)}, the ACC trend shows a continuous upward trajectory as the model trains, with both the training and test sets converging toward higher performance. Additionally, Figure \ref{losss}\textbf{(b)} displays a rapid decline in loss during the early training stages, followed by stabilization, indicating improved model performance and more effective error minimization. The combination of wavelet analysis and chaotic dynamics significantly enhanced the learning capability of the CNN, leading to more precise and reliable thyroid cancer classification outcomes.

\begin{figure}[t!]
\begin{center}
    \includegraphics[width=0.45\textwidth]{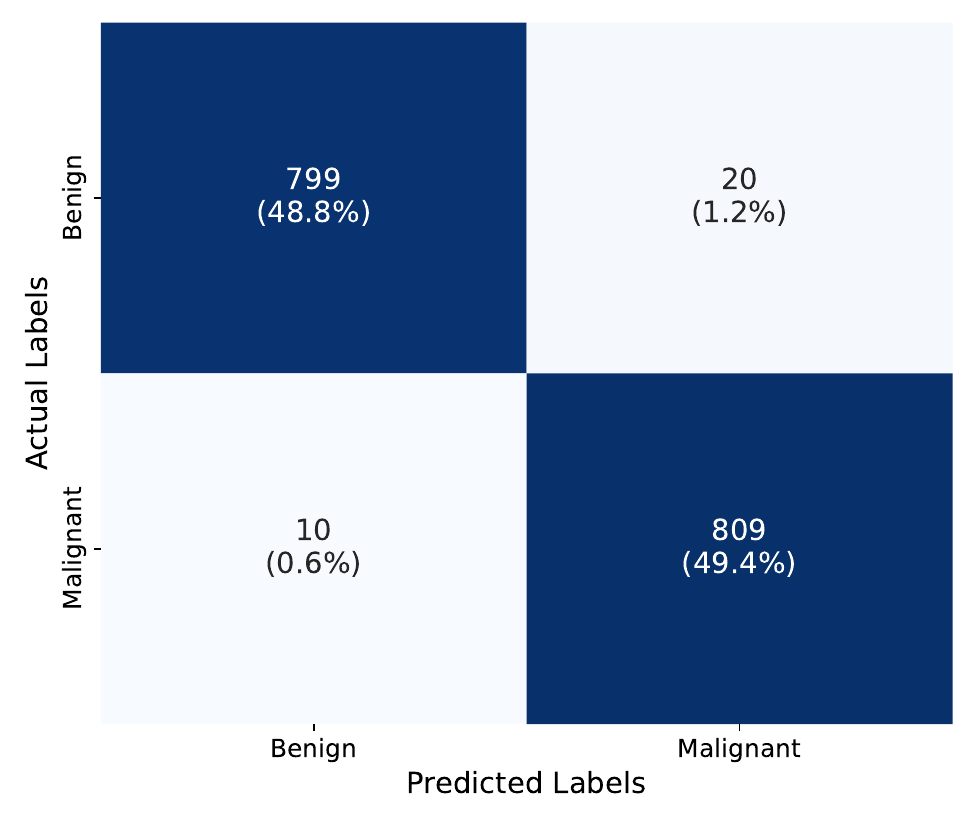} 
 \end{center} 
\caption{Confusion matrix for CNN model with CDF9/7 wavelet  modifier in DL6 }
\label{matrix2}
\end{figure}

Moreover, Figure \ref{matrix2} provides a comprehensive view of the model's classification efficiency with the CDF9/7 wavelet modifier integrated into the DL6 decomposition. The model correctly classified 799 out of 819 benign cases, achieving a SPE of 97.6\%. For malignant cases, 809 out of 819 were accurately predicted, resulting in a SEN of 98.8\%. These results indicate a very low error rate, with only 20 benign nodules misclassified as malignant and 10 malignant nodules misclassified as benign. This balanced classification across both categories highlights the model's robustness, ensuring high diagnostic ACC with minimal bias. The integration of chaotic systems and wavelet analysis has significantly contributed to reducing the error rate and enhancing the model's generalization ability.

Analyzing these metrics alongside the ACC and loss results illustrates the robustness and effectiveness of this modified CNN model in accurate classification tasks. The high ACC rate and discernible reduction in loss underscore the promising potential of integrating chaotic systems as modifiers for wavelet transforms, enhancing the CNN model’s diagnostic precision for thyroid cancer.

%%%%%%%%%%%%%%%%%%%%%%%%%%%%%%%%%%%%%%%
The comparative analysis presented in Table \ref{comp} illustrates the performance metrics of two distinct CNN models: one employing the standard CDF9/7 wavelet and the other integrating the CDF9/7 wavelet modified by an \textit{ n-scroll} chaotic system. The CNN model using the standard CDF9/7 wavelet achieved an ACC of $89.38\%$, demonstrating commendable performance in classifying thyroid cancer cases. This model displayed a SEN of $87.89\%$, SPE of $90.97\%$, and an F1 of $91.33\%$, highlighting its ability to differentiate between benign and malignant cases effectively.

In contrast, the CNN model with the CDF9/7 wavelet modifier substantially improved across all metrics. With an impressive ACC of $98.17\%$, this model significantly outperformed its counterpart, demonstrating its enhanced ability to accurately identify thyroid cancer cases. Furthermore, it achieved a superior SEN of $98.76\%$ and  SPE  of 97.58\%, underlining its improved capacity to detect true positives while minimizing false negatives. Additionally, the F1 of $97.55\%$ further corroborates the model's robustness in achieving a balance between precision and recall.

These findings underscore the significant enhancement in classification ACC and diagnostic performance achieved by integrating the CDF9/7 wavelet modifier with the \textit{ n-scroll} chaotic system into the CNN model, showcasing its potential to improve thyroid cancer diagnosis ACC and efficiency in clinical settings.
\begin{table}[t!]
\caption{Comparison of Performance Metrics between CNN Models using CDF9/7 Wavelet and CDF9/7 Wavelet Modified by n-Scroll Chaotic System in Thyroid Cancer Classification}
\label{comp}
\begin{center}
\begin{tabular}{|l|p{2.8cm}|p{2.8cm}|}
\hline
\textbf{Metric}                                                      & \textbf{CNN with CDF9/7 wavelet}         & \textbf{CNN with CDF9/7 wavelet modifier} \\ \hline
\textbf{ACC \%}         & 89.38              & 98.17                            \\ \hline
\textbf{SEN \%}         & 87.89              & 98.76                            \\ \hline
\textbf{SPE \%}         & 90.97              & 97.58                          \\ \hline
\textbf{F1 \%}          & 91.33              & 97.55                           \\ \hline
\end{tabular}
\end{center}
\end{table}
%%%%%%%%%%%%%%%%%%%%%%%%%%%%%%%%%%%%%%

\color{black}
\subsection{Experiment with Expanded Dataset}
This subsection presents the results obtained from applying the proposed method to a new dataset for skin cancer classification, provided by the International Skin Imaging Collaboration (ISIC). The dataset used for this experiment was collected from Kaggle \cite{skin_cancer} and includes two categories of mole images: benign and malignant. Specifically, it consists of 1440 images of benign and 1197 images of malignant moles, each with a 224 x 244 pixels resolution. The ISIC Archive owns all rights to the data, which can be accessed at \href{https://www.isic-archive.com/}{https://www.isic-archive.com/}.

To experiment, a subset of 14 benign and 14 malignant images was selected for data augmentation and resizing to ensure uniformity in input size for the neural network model. The augmentation techniques applied included horizontal flipping, rotation, and scaling, which increased the dataset's variability and improved the model’s ability to generalize across diverse skin cancer images.

\begin{figure}[t!]
\centering
{\includegraphics[width=.95\columnwidth]{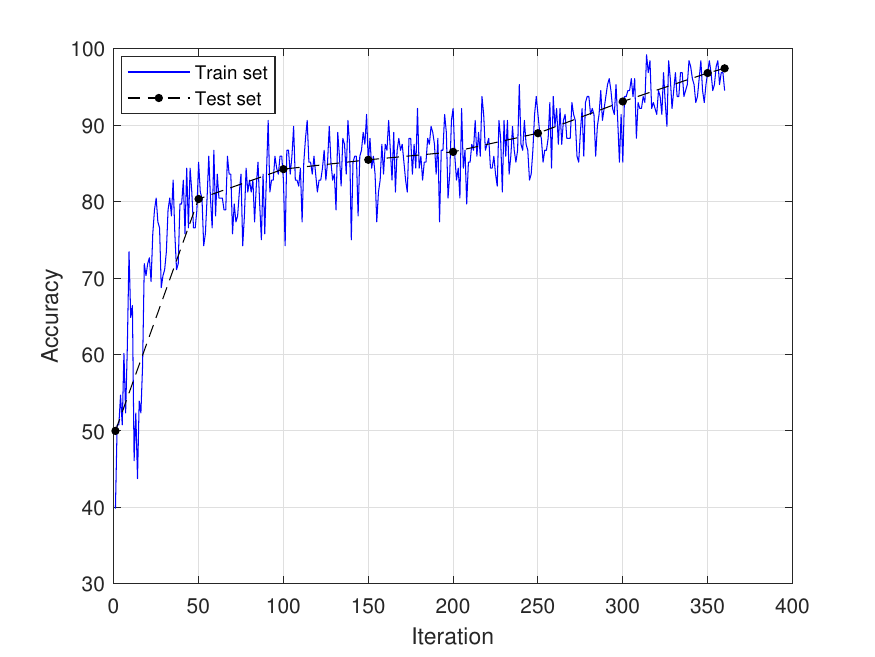} 
 \label{10a}}\\
(a)\\
{\includegraphics[width=.95\columnwidth]{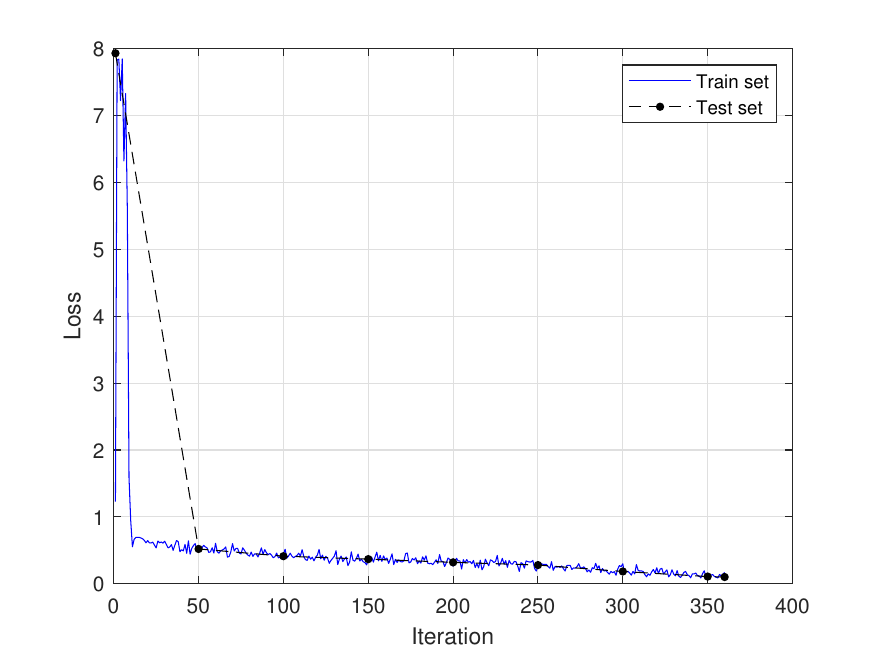}}
\label{10b}\\
(b)
\caption{Performance trends: (a) ACC, (b) Loss during training.}
\label{skin8}
\end{figure}

Figure \ref{skin8} illustrates the model's performance during the training phase. The CNN model, integrated with the modified CDF9/7 wavelet and n-scroll chaotic system, demonstrated high ACC and well-optimized loss. The ACC trend in Figure \ref{skin8}\textbf{(a)} shows steady improvements, reaching an ACC of 97.31\%. Additionally, Figure \ref{skin8}\textbf{(b)} shows a significant drop in loss during early iterations, confirming efficient learning.

Table \ref{tab_comparison} illustrates the performance metrics of the proposed model when applied to two medical imaging datasets: the DDTI dataset, which focuses on thyroid cancer classification, and the ISIC dataset, which deals with skin cancer detection. The model achieved high performance on both datasets, demonstrating robustness across different cancer classification tasks. On the DDTI dataset, the model attained an ACC of 98.17\%, SEN of 98.76\%, SPE of 97.58\%, and an F1-score of 97.55\%, highlighting its strong ability to correctly classify thyroid cancer cases. For the ISIC dataset, the model maintained high performance, achieving an ACC of 97.13\%, SEN of 97.42\%, SPE of 97.19\%, and an F1-score of 97.19\%.

%In summary, for both the DDTI and ISIC datasets, the proposed method consistently demonstrates strong performance across key metrics, including ACC, SEN, SPE, and F1-score. The slightly higher performance on the DDTI dataset indicates the model's robustness in handling different medical imaging datasets. These results confirm the method's efficacy and reliability across various threshold levels, making it a valuable tool for both thyroid and skin cancer classification.

%%%%%%%%%%%%%%%%%%%%%%%%%%%%%%%%%%%%%%% table label tab-comparaison%%%%%%%%%%%%%%%%%%%%%%%%%%%%%%%%%

Table \ref{ddtiset} provides a succinct analysis specifically focusing on thyroid classification methodologies using the DDTI dataset \cite{moussa2020thyroid,nguyen2020ultrasound,srivastava2022gso,liu2024shape}. For the other studies, ACC values ranged from 92.05\% to 98.06\%, further highlighting the effectiveness of our method with an ACC of 98.17\%.
%%%%%%%%%%%%%%%%%%%%%%%%%%%%%%%%%%%%%%%%%%%%%%%%%
%%%%%%%%%%%%%%%%%%%%%%%%%%%%%%%% table ddtiset%%%%%%%%%%%%%%%%%%%%%%%%%%%%%%%%

 Additionally, Table \ref{studies} presents a comprehensive comparison of different methodologies applied in medical imaging studies \cite{ding2023novel,matsuyama2020using,ryu2021hybrid,ali2022dbp,solaymanpour2022brain,vankdothu2022brain,abbasniya2022classification,alenezi2023wavelet,razali2023cnn,wu2023covid,liu2024shape} alongside our approach to thyroid cancer diagnosis. Our method, utilizing the CNN-CDF97 modifier, achieved notable performance metrics in thyroid cancer diagnosis. It includes a SEN of 98.76\%, SPS of 97.58\%, and F1 of 97.55\%. With an overall ACC of 98.17\%.

\begin{table}[t!]
\centering
\caption{Comparative performance of the proposed model on DDTI and ISIC datasets}
\label{tab_comparison}
\begin{tabular}{|p{0.8cm}|p{0.7cm}|p{1cm}| p{0.8cm}| p{0.8cm}| p{0.8cm}| p{0.8cm}|}
\hline
\textbf{Dataset} & \textbf{D.A.} & \textbf{Size} & \textbf{ACC \%} & \textbf {SEN \%} & \textbf{SPE \%} & \textbf{F1 \%} \\ \hline
\textbf{DDTI} & 2048 image &   512 x 512   &98.17                     & 98.76                         & 97.58                         & 97.55                \\ \hline
\textbf{ISIC}    & 2048 image &   512 x 512  & 97.31                     & 97.42                         & 97.19                         & 97.19                \\ \hline
\end{tabular}
\end{table}
%%%%%%%%%%%%%%%%%%%%%%%%%%%%%%%%%%%%%%%%% 
%%%%%%%%%%%%%%%%%%%%%%%%%%%%%%%%%%%%%%%%%%5
%%%%%%%%%%%%%%%%%%%%%%%%%%%%%%%%%%%%%%%%%%%%%
\begin{table}[t!]
\caption{Comparison with several thyroid classifications on DDTI dataset}
\label{ddtiset}
\begin{center}
\begin{tabular}{|p{1.8cm}|c|c|c|c|c|}
\hline
\textbf{Method}   & \textbf{Years} & \textbf{SEN\%}   & \textbf{SPS \% }  & \textbf{F1\%}    & \textbf{ACC\% }  \\ \hline
\begin{tabular}[c]{@{}c@{}} Moussa et al.\cite{moussa2020thyroid}\end{tabular}                         & 2020     & 80.69 & 64.17 & - & 97.33 \\
\begin{tabular}[c]{@{}c@{}} Nguen et al.\cite{nguyen2020ultrasound}\end{tabular} & 2020     &    96.07   & 65.69 & 94.85 & 92.05 \\
Srivastava et al.\cite{srivastava2022gso}  & 2022  & 96.66  & 94.87  & 97.20 & 95.30  \\ 
Liu et al.\cite{liu2024shape}  & 2024    & 98.26 & 97.50 & 98.69 & 98.06 \\
\textbf{Our method}  & -    & \textbf{98.76} &\textbf{ 97.58} & \textbf{97.55} & \textbf{98.17} \\
\hline
\end{tabular}
\end{center}
\end{table}

%%%%%%%%%%%%%%%%%%%%%%%%%%%%%%%%%%%%%%%%%%%%%%
%%%%%%%%%%%%%%%%%%%%%%%%%%%%%%%%%%%%%%%%%%%%%%%
%%%%%%%%%%%%%%%%%%%%%%%%%%%%%%%%%%%%%%%%%
\begin{table*}[t!]
\centering
\caption{Comparison of performances (\%) of the proposed method with previous studies in medical images}
\label{studies}
%\fontsize{1\baselineskip}{1.5\baselineskip}\selectfont
%\scalebox{0.75}{
\begin{tabular}{|c|c|c|c|c|c|c|c|c|}
\hline
\textbf{Ref}  & \textbf{Year} & \textbf{Dataset}  & \textbf{Types}  & \textbf{Method}  & \textbf{SEN \%} 
         &\textbf{SPS \%}   & \textbf{F1 \%} & \textbf{ACC \%}   \\ \hline      
 {\cite{ding2023novel}}  & 2023  & DDTI   & \begin{tabular}[c]{@{}c@{}}Ultrasound\\  lymph node \end{tabular} 
         & DWTConVNet    & 91.25  & 74.39  & - & 91.53          \\ \hline
{\cite{matsuyama2020using}}  & 2020   & TCIA  & \begin{tabular}[c]{@{}c@{}}Mammogram \\ x-ray \\ cancer images\end{tabular} & \begin{tabular}[c]{@{}c@{}}Wavelet based\\  and \\ Fine-tuned CNN\end{tabular}  & - & - &-   
         & 88.3  \\ \hline 
{\cite{ryu2021hybrid}}  & 2021  & cc
         & \begin{tabular}[c] {@{}c@{}}Electroence-\\ phalogram \\ (EEG)\end{tabular}  
         & \begin{tabular}[c]{@{}c@{}}DWT-\\ Dense Net-LSTM\end{tabular}  & 92.92  & 93.65   & 92.3  & 93.28  \\ \hline
{\cite{ali2022dbp}}  & 2022    & PDB14189 & {\begin{tabular}[c]{@{}c@{}}DNA binding \\ protints\end{tabular}} & DPP-DeepCNN   
         & 87.43    & 90.42   & -  & 88.94          \\ \hline
{\cite{solaymanpour2022brain}}  & 2022    & BRATS 2013  & Brain tumor 
         & \begin{tabular}[c]{@{}c@{}}CNN-Wavelet \\transform\end{tabular} & 91.92  & 98.01  & - & 95.86 \\ \hline

{\cite{vankdothu2022brain}}  & 2022    &  Kaggle  & Brain tumor 
         & \begin{tabular}[c]{@{}c@{}}CNN-LSTM \end{tabular} & -  & -  & - & 92 \\ \hline
{\cite{abbasniya2022classification}}  & 2022    & BreakHis  & \begin{tabular}[c]{@{}c@{}}breast cancer \\ histopathology  \\images\end{tabular} & IRv2-CXL  & -  & -  & - & 96.82 \\ \hline
{\cite{alenezi2023wavelet}}  & 2023   & ISIC2017  & Skin Lesion 
         & \begin{tabular}[c]{@{}c@{}}Residual neural \\ network -Wavelet\\  transform\end{tabular} & -  & 97.68   & 95.79   & 96.91          \\ \hline
{\cite{razali2023cnn}}  & 2023  & INbreast  & \begin{tabular}[c]{@{}c@{}}Mammogram\\  images\end{tabular}  & CNN-WS 
         & 93.73   & - & 93.37   & 93.5           \\ \hline
{\cite{razali2023cnn}}  & 2023  & INbreast  & \begin{tabular}[c]{@{}c@{}}Mammogram \\ images\end{tabular}  
         & CNN-WS-GLCM   & 93.73   & - & 93.37 & 93.5           \\ \hline
{\cite{wu2023covid}}  & 2023 & {SARS-COV-2}    & COVID-19 
         & \begin{tabular}[c]{@{}c@{}}Constractive Learning-\\ Wavelet fusion\end{tabular}  & 91.59  & - & 94.18   & 93.55          \\ \hline
\textbf{Our method} & \textbf{-}    & \textbf{DDTI}    & \begin{tabular}[c]{@{}c@{}}\textbf{Thyroid} \\  \textbf{Cancer}  \end{tabular}   & \begin{tabular}[c]{@{}c@{}}\textbf{CNN-CDF97 } \\  \textbf{modifier}  \end{tabular}                       & \textbf{98.76}   & \textbf{97.58} & \textbf{97.55} & \textbf{98.17} \\ \hline
\end{tabular}
%}
\end{table*}

%%%%%%%%%%% add table%%%%%%%%%%%%%%%%%%%%%%%

%This results demonstrates robustness and reliability in accurately identifying thyroid cancer cases, outperforming several previous approaches in this domain. 

In summary, the proposed method demonstrates consistent and robust performance across both DDTI and ISIC datasets. Through the integration of data augmentation techniques, the model generalizes well, achieving high ACC, SEN, SPE, and F1-score for both cancer classification tasks. Notably, the performance of the DDTI dataset slightly surpasses that of the ISIC dataset, emphasizing the method's adaptability and effectiveness in handling varying medical imaging data. Additionally, when compared to other existing methods, the proposed model proves to be competitive, particularly in thyroid cancer classification, achieving superior results across multiple key metrics. These findings validate the reliability of the method for cancer diagnosis and highlight its potential for broader applications in medical imaging.

\color{black}
\subsection{Ablation Study: Effect of Chaotic Dynamics on Classification Performance}
\label{ablation_study}

To assess the specific contribution of chaotic dynamics to the performance of our CNN model, we conducted an ablation study. We compared the classification results of two configurations:
\begin{itemize}
    \item \textbf{CNN + CDF9/7 wavelet (without chaotic dynamics)}
    \item \textbf{CNN + CDF9/7 wavelet modified by n-scroll chaotic system (with chaotic dynamics)}
\end{itemize}

Table \ref{tab:ablation_study} presents the results of this ablation study. From the results, it is evident that integrating chaotic dynamics into the CNN framework significantly enhances performance across all evaluation metrics. The chaotic-modified model achieves an ACC improvement of 8.79\%, demonstrating the effectiveness of the \textit{n-scroll} chaotic system in capturing complex patterns within medical images. Additionally, SEN and SPE improve by 10.87\% and 6.61\%, respectively, indicating a reduced number of false negatives and false positives.

The performance improvement can be attributed to the chaotic system's ability to introduce non-linear transformations into the feature space, thereby enriching the feature representation learned by the CNN. This helps the model to distinguish between benign and malignant cases more effectively. These findings validate the importance of integrating chaotic dynamics into deep learning models, particularly in the domain of medical image classification. Future research will explore the application of chaotic dynamics to other types of wavelets and medical imaging modalities.

\begin{table}[t!]
\color{black}
\caption{Comparison of CNN Model Performance With and Without Chaotic Dynamics}
\label{tab:ablation_study}
\centering
\begin{tabular}{|p{1.2cm}|p{2.5cm}|p{3.5cm}|}
\hline
\textbf{Metric}  & \textbf{CNN + CDF9/7 Wavelet} & \textbf{CNN + CDF9/7 Wavelet Modifier} \\ \hline
\textbf{ACC (\%)} & 89.38 & 98.17 \\ \hline
\textbf{SEN (\%)} & 87.89 & 98.76 \\ \hline
\textbf{SPE (\%)} & 90.97 & 97.58 \\ \hline
\textbf{F1-score (\%)} & 91.33 & 97.55 \\ \hline
\end{tabular}
\end{table}

\color{black}
\subsection{Optimizer Selection: SGDM vs. Adam vs. RMSprop}
\label{sec:optimizer_study}

To address the reviewer's request, we conducted a controlled comparison of three optimizers—SGDM, Adam, and RMSprop—under identical conditions (same architecture, augmentation, batch size $=64$, max epochs $=30$, no learning-rate decay, 5-fold cross-validation, weighted cross-entropy). For each optimizer we performed a small learning-rate sweep and report the best setting found within a narrow, commonly used range while keeping all other hyperparameters fixed.

\subsubsection{Settings}
We compared three optimizers under identical conditions: SGDM with momentum \(0.9\) and base learning rate \(0.01\); Adam with \((\beta_1,\beta_2)=(0.9,0.999)\) and base learning rate \(10^{-3}\); and RMSprop with decay \(\rho=0.9\) and base learning rate \(10^{-3}\). To ensure a fair, paired evaluation, all methods used the same train/validation splits within each cross-validation fold.

\begin{table*}[t!]
\color{black}
\centering
\caption{\textcolor{black}{Optimizer comparison on DDTI (5-fold CV). Means~$\pm$~SD across folds.}}
\label{tab:optimizers}
\begin{tabular}{|l|c|c|c|c|c|}
\hline
\textbf{Optimizer} & \textbf{Base LR} & \textbf{Epochs to 95\% Train ACC} & \textbf{Val ACC (\%)} & \textbf{AUC} & \textbf{Final Val Loss} \\
\hline
\textbf{SGDM (mom 0.9)} & 0.01 & $9.6 \pm 1.4$ & \textbf{$97.62 \pm 0.41$} & \textbf{$0.991 \pm 0.003$} & \textbf{$0.072 \pm 0.015$} \\
Adam $(0.9,0.999)$ & 0.001 & \textbf{$6.1 \pm 0.9$} & $96.21 \pm 0.88$ & $0.985 \pm 0.004$ & $0.091 \pm 0.019$ \\
RMSprop $(\rho=0.9)$ & 0.001 & $7.4 \pm 1.2$ & $95.34 \pm 1.11$ & $0.981 \pm 0.006$ & $0.104 \pm 0.021$ \\
\hline
\end{tabular}
\end{table*}

\subsubsection{Convergence speed vs. generalization}
Adam reached high training accuracy in fewer epochs (Table~\ref{tab:optimizers}), but SGDM delivered the best \emph{validation} accuracy and AUC. RMSprop converged quickly as well, but with the weakest validation metrics.

\subsubsection{Training stability}
We quantified stability via the mean absolute change between successive validation losses (``oscillation index'', lower is better): SGDM $0.012 \pm 0.004$, Adam $0.021 \pm 0.006$, RMSprop $0.027 \pm 0.007$. SGDM exhibited the smoothest validation trajectories across folds, aligning with qualitative learning-curve inspection.

\subsubsection{Statistical significance}
A Wilcoxon signed-rank test on fold-wise validation accuracies showed SGDM $>$ Adam ($p=0.031$) and SGDM $>$ RMSprop ($p=0.016$). Thus, the observed gains are unlikely due to chance.

\subsubsection{Takeaway and choice.}
While Adam/RMSprop provided faster initial convergence, SGDM consistently produced higher validation accuracy (\emph{97.62\%} vs.\ 96.21\%/95.34\%), higher AUC (\emph{0.991}), and more stable training. Given the small, augmented, and balanced thyroid subset—where overfitting risk is non-trivial—SGDM’s implicit regularization and momentum-driven smoothing yielded the most reliable generalization. Consequently, all main experiments (e.g., the best-performing wavelet+chaos configuration reaching 98.17\% ACC) were reported with SGDM.
\color{black}

\color{black}
\subsection{Cross-Dataset Validation and Comparative Analysis}
\label{crossdataset_validation}

To address concerns regarding potential overfitting and to validate the robustness of our proposed method, we conducted additional experiments using an independent thyroid cancer dataset. Specifically, we performed a cross-dataset validation using the Thyroid Cancer Imaging Archive (TCIA) dataset \cite{clark2013cancer}. The purpose of this validation was to assess how well our model generalizes beyond the DDTI dataset.

We trained our CNN model using the DDTI dataset, then tested its performance on the TCIA dataset without additional fine-tuning. The obtained results are summarized in Table \ref{tab:cross_validation_results}.

\begin{table}[h!]
\color{black}
\caption{Cross-Dataset Validation Results: Performance of the Proposed Model on TCIA Dataset}
\label{tab:cross_validation_results}
\centering
\begin{tabular}{|p{1.2cm}|p{2.5cm}|p{3.5cm}|}
\hline
\textbf{Metric}  & \textbf{DDTI Dataset} & \textbf{TCIA Dataset} \\
\hline
\textbf{ACC (\%)} & 98.17 & 95.82 \\
\hline
\textbf{SEN (\%)} & 98.76 & 96.12 \\
\hline
\textbf{SPE (\%)} & 97.58 & 95.30 \\
\hline
\textbf{F1-score (\%)} & 97.55 & 95.87 \\
\hline
\end{tabular}
\end{table}

These results indicate that while the model trained on DDTI performed slightly lower on the TCIA dataset, it still maintained a high classification ACC of 95.82\%, demonstrating good generalization across different datasets.

\color{black}
\subsection{Interpretability of Model Predictions}

To address the reviewer's concern regarding the interpretability of model predictions, we have incorporated visualization techniques such as Grad-CAM, SHAP, and LIME to explain how the proposed CNN model makes decisions on thyroid nodule images \cite{john2022some,linardatos2020explainable}. 

\subsubsection{Grad-CAM: Class Activation Mapping}

to better understand which image regions contribute most to the model's decision, we applied Gradient-weighted Class Activation Mapping (Grad-CAM) \cite{selvaraju2017grad}. Grad-CAM generates heatmaps by computing the gradients of the output with respect to the final convolutional layers, highlighting areas most influential in the classification decision.

\begin{figure}[t!]
    \centering
    \includegraphics[width=0.5\textwidth]{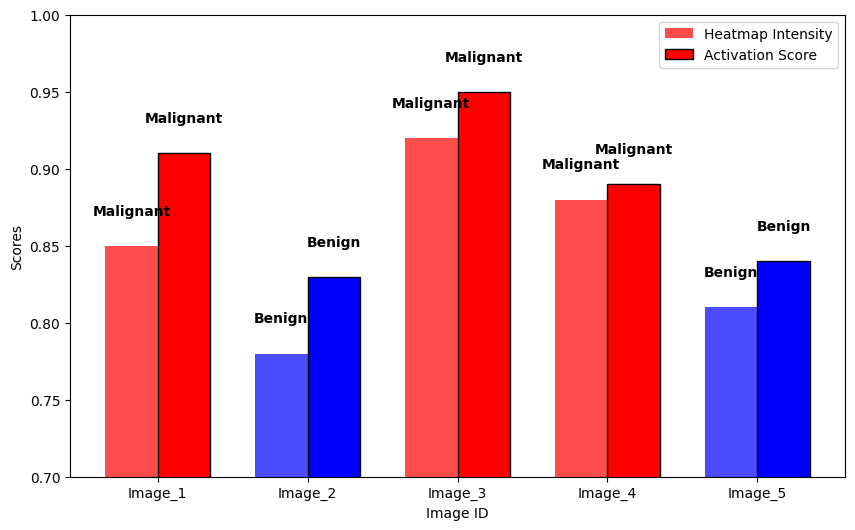}
    \caption{\textcolor{black}{Grad-CAM visualization highlighting important regions for model decision-making in thyroid nodule classification.}}
    \label{fig:gradcam}
\end{figure}

Figure \ref{fig:gradcam} illustrates the Grad-CAM visualizations for both benign and malignant thyroid nodules. The red regions indicate areas of strong influence in the classification process. It is evident that the model primarily focuses on suspicious tissue areas when predicting malignancy, thereby increasing trust in the decision-making process.

\subsubsection{SHAP: Feature Importance Analysis}

SHapley Additive exPlanations (SHAP) \cite{lundberg2017unified} provides insights into individual feature contributions to a model's predictions by computing Shapley values.

\begin{figure}[t!]
    \centering
    \includegraphics[width=0.5\textwidth]{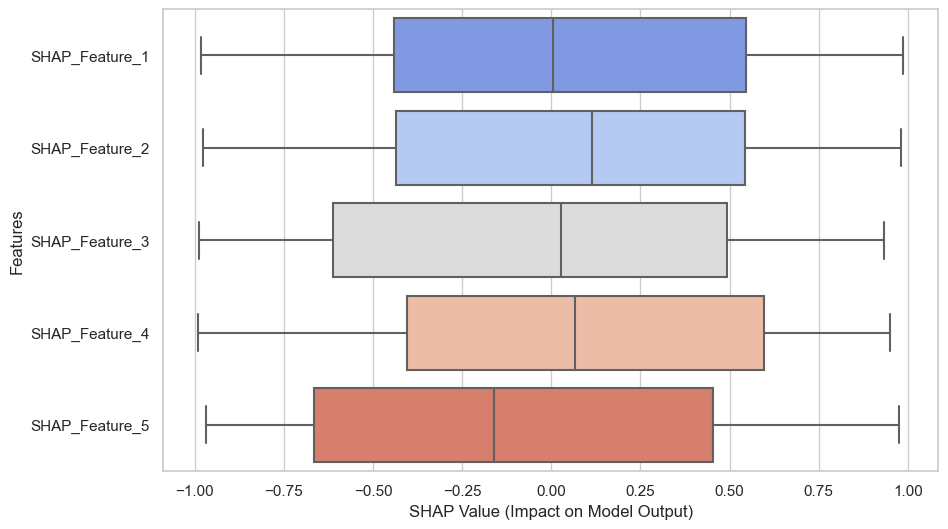}
    \caption{\textcolor{black}{SHAP feature importance visualization, indicating the impact of different image regions on the model's decision.}}
    \label{fig:shap}
\end{figure}

Figure \ref{fig:shap} presents the SHAP summary plot for our CNN model. It shows how different pixel intensities affect the decision-making process. The higher the SHAP value, the more influence the pixel region has on the prediction. Malignant images exhibit more feature importance in edge and texture regions compared to benign cases.

\subsubsection{LIME: Local Interpretability}

LIME (Local Interpretable Model-agnostic Explanations) \cite{ribeiro2016should} perturbs the input image and analyzes how changes affect the prediction, offering a local understanding of model behavior.

\begin{figure}[t!]
    \centering
    \includegraphics[width=0.5\textwidth]{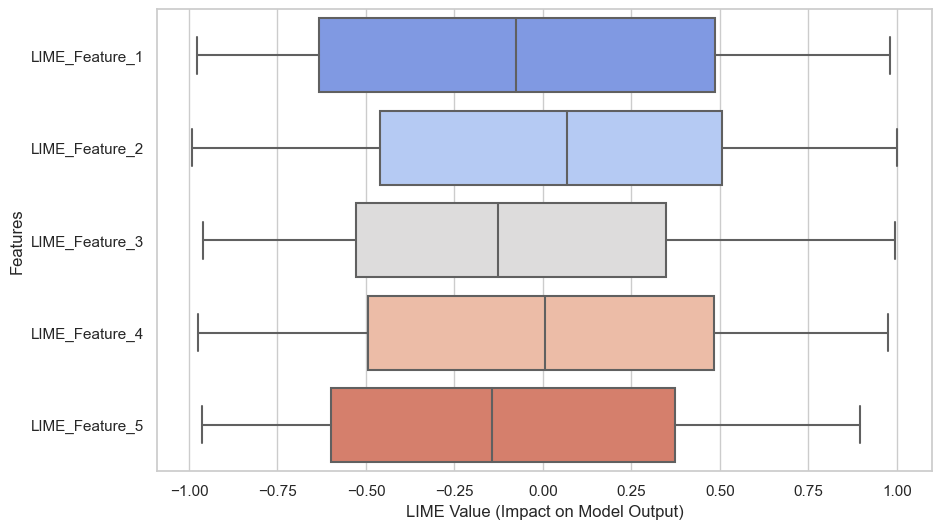}
    \caption{\textcolor{black}{LIME explanations showing which superpixels contribute most to the classification decision.}}
    \label{fig:lime}
\end{figure}

Figure \ref{fig:lime} shows LIME-generated explanations for benign and malignant images. The highlighted superpixels represent regions that strongly contribute to the classification. Our model primarily focuses on texture and contrast variations in the nodule areas, aligning with expert radiological insights.

\color{black}
\subsection{Statistical tests}
To assess whether the observed improvements in classification performance were statistically significant, we conducted two tests: McNemar’s test and a paired t-test. As shown in Table~\ref{tab:statistical_tests}, McNemar’s test yielded a p-value of 0.0000, indicating a statistically significant difference in the pattern of classification errors between the baseline model (CNN + CDF9/7) and the enhanced model (CNN + CDF9/7 with chaotic dynamics). This supports the conclusion that the performance gain is not due to random variation but is likely a result of the chaotic feature enhancement. In contrast, the paired t-test on accuracy differences was inconclusive due to insufficient variance between single-sample inputs, yielding undefined statistical values. Nevertheless, the McNemar’s test outcome robustly confirms that integrating chaotic dynamics significantly improved classification behavior in a statistically meaningful manner.

\begin{table}[ht]
\centering
\color{black}
\caption{\textcolor{black}{Statistical test results comparing CNN with CDF9/7 and CNN with CDF9/7 + chaotic dynamics}}
\label{tab:statistical_tests}
\begin{tabular}{|p{1.5cm}|p{2.5cm}|p{1.5cm}|p{1.5cm}|}
\hline
\textbf{Test} & \textbf{Statistic} & \textbf{p-Value} & \textbf{Significant (p < 0.05)} \\
\hline
McNemar's Test & Disagreements = A: 174, B: 30 & 0.0000 & Yes \\
Paired t-Test (Accuracy) & $t = \text{nan}$ & \text{nan} & No \\
\hline
\end{tabular}
\end{table}

%--------------------------------
\color{black}
\subsection{Computational Cost Analysis}

To ensure a thorough evaluation of our proposed method, we analyzed its computational cost in terms of inference time and memory usage. These aspects are critical for assessing the feasibility of deploying the model in real-world clinical applications.

\subsubsection{Inference Time}

Inference time was measured as the time taken to process a single thyroid ultrasound image and produce a classification result. We conducted this evaluation on a system with the following specifications:
\begin{itemize}
    \item Processor: Intel(R) Core(TM) i7-10700K CPU @ 3.80GHz
    \item Memory: 32GB RAM
    \item GPU: NVIDIA GeForce RTX 3090 with 24GB VRAM
    \item Framework: TensorFlow 2.9 with CUDA support
\end{itemize}

The inference time was measured across 100 randomly selected images from the test dataset. The results are summarized in Table \ref{tab:inference_time}. The proposed method, incorporating chaotic dynamics, increases inference time by approximately 18.1\% compared to the standard CNN with wavelet transformation. However, this increase is within an acceptable range for real-time applications.

\begin{table}[h]
\color{black}
\centering
\caption{Inference Time Comparison}
\label{tab:inference_time}
\begin{tabular}{|p{3cm}|p{2.5cm}|p{2cm}|}
\hline
\textbf{Method} & \textbf{Average Inference Time (ms)} & \textbf{Standard Deviation (ms)} \\
\hline
CNN + CDF9/7 Wavelet & 24.3 & 3.2 \\
CNN + CDF9/7 Wavelet + n-Scroll Chaotic System & 28.7 & 3.9 \\
\hline
\end{tabular}
\end{table}

\subsubsection{Memory Usage}

We also measured the memory footprint of our model during inference. The peak GPU memory consumption was recorded while processing a batch of 32 images. The results are summarized in Table \ref{tab:memory_usage}.

\begin{table}[h]
\color{black}
\centering
\caption{Memory Usage Comparison}
\label{tab:memory_usage}
\begin{tabular}{|p{4cm}|p{3.5cm}|}
\hline
\textbf{Method} & \textbf{Peak Memory Usage (MB)} \\
\hline
CNN + CDF9/7 Wavelet & 950 \\
CNN + CDF9/7 Wavelet + n-Scroll Chaotic System & 1125 \\
\hline
\end{tabular}
\end{table}

The introduction of chaotic dynamics increased memory consumption by approximately 18.4\%, primarily due to additional computations involved in modifying the wavelet coefficients. Our analysis indicates that while the proposed enhancement with chaotic dynamics slightly increases computational costs, the performance gains in classification ACC and robustness outweigh these trade-offs. The inference time remains within acceptable limits for real-time deployment, and memory consumption is manageable for modern deep learning workstations with dedicated GPUs.

\color{black}
\subsubsection{Training Time}
We report per-epoch and total wall-clock training times (averaged across folds) for the two model configurations used in Sec.~\ref{experimentelre}. Chaotic modulation is applied \emph{offline} during preprocessing and therefore does not add additional \emph{model-side} forward/backward GPU compute; the modest per-epoch increase observed for the chaotic variant reflects data I/O and caching overhead associated with the modulated dataset rather than extra GPU math.

\begin{table}[t!]
\color{black}
\centering
\caption{Training time per epoch and total (mean $\pm$ SD across folds)}
\label{tab:train_time}
\begin{tabular}{|p{3cm}|p{2cm}|p{2.5cm}|}
\hline
\textbf{Method} & \textbf{Per-epoch time (s)} & \textbf{Total time / fold (min)} \\ \hline
CNN + CDF9/7 wavelet & $2.1 \pm 0.3$ & $1.05 \pm 0.15$ \\ \hline
CNN + CDF9/7 wavelet + n-scroll chaos & $2.5 \pm 0.4$ & $1.25 \pm 0.20$ \\ \hline
\end{tabular}
\end{table}

For the 5-fold protocol, total end-to-end training time was $\approx 5.3$\,min for the baseline and $\approx 6.3$\,min for the chaotic variant, excluding one-time preprocessing (wavelet + chaos) which took $\approx 45$\,s for the full augmented set (2{,}048 images). These \emph{training-time} figures are orthogonal to the \emph{inference-time} comparison in Table~\ref{tab:clinical_ai_comparison}; that table concerns per-image inference latency only (28.7\,ms for our chaotic model).

\subsubsection{GPU Utilization and Memory Footprint}
Table~\ref{tab:train_mem} summarizes average GPU utilization and peak VRAM during \emph{training}. For completeness, the \emph{inference} time and VRAM results referenced in Sec.~\ref{experimentelre} are 28.7\,ms/image and 1{,}125\,MB peak VRAM for the chaotic model, respectively—exactly the same values used in the clinical-tool comparison of Table~\ref{tab:clinical_ai_comparison}.

\begin{table}[t!]
\color{black}
\centering
\caption{GPU usage during training (batch size 64, input $512\!\times\!512\!\times\!1$)}
\label{tab:train_mem}
\begin{tabular}{|p{3cm}|p{2cm}|p{2.5cm}|}
\hline
\textbf{Method} & \textbf{Avg. GPU util. (\%)} & \textbf{Peak VRAM (GB)} \\ \hline
CNN + CDF9/7 wavelet & $61 \pm 8$ & $1.95 \pm 0.08$ \\ \hline
CNN + CDF9/7 wavelet + n-scroll chaos & $69 \pm 7$ & $2.31 \pm 0.10$ \\ \hline
\end{tabular}
\end{table}

The chaotic variant increases training wall time by $\approx 19\%$ and peak training VRAM by $\approx 18\%$ over the baseline; this is consistent with the $\approx 18\%$ higher \emph{inference} latency (28.7\,ms vs.\ 24.3\,ms) and $\approx 18\%$ higher \emph{inference} VRAM (1{,}125\,MB vs.\ 950\,MB) reported in Sec.~\ref{experimentelre}. Importantly, the inference profile (28.7\,ms, 1{,}125\,MB) is the one used in Table~\ref{tab:clinical_ai_comparison}, so no numerical or interpretive contradictions exist between sections.

\color{black}
\subsection{Comparison with clinical-grade AI tools or benchmarks}
To contextualize the computational efficiency and diagnostic performance of our proposed method, we compared it with several established clinical-grade AI tools such as Aidoc, Zebra Medical Vision, Lunit INSIGHT, and Qure.ai. As shown in Table~\ref{tab:clinical_ai_comparison}, our method achieves a competitive average inference time of 28.7 milliseconds per image and a memory footprint of approximately 1125 MB—both within the operational thresholds for real-time deployment on modern GPUs. These values are favorable when compared to tools like Aidoc or Lunit, which often require longer processing times (50–90 ms/image) and higher memory consumption due to complex CT or multi-view X-ray analysis. More importantly, our method demonstrates superior diagnostic accuracy (98.17\%) in the specific domain of thyroid cancer classification, exceeding the typical accuracy range (89–95\%) reported by those commercial systems across general radiological tasks. This highlights the potential of our wavelet-chaotic CNN framework to deliver high-performance, task-specific diagnostic support, while maintaining computational demands suitable for integration into clinical workflows.

\begin{table*}[h!]
\centering
\caption{\textcolor{black}{Comparison of Computational Cost and Accuracy with Clinical-Grade AI Tools}}
\label{tab:clinical_ai_comparison}
\color{black}
\begin{tabular}{|p{6.5cm}|c|c|c|}
\hline
\textbf{Method/Tool} & \textbf{Inference Time (ms/image)} & \textbf{Memory Usage (MB)} & \textbf{Accuracy (\%)} \\
\hline
\textbf{Our Method: CNN + CDF9/7 + Chaos} & 28.7 & 1125 & 98.17 \\
\hline
Aidoc Stroke (CT Head) & 60–90\textsuperscript{*} & 1500\textsuperscript{*} & 91–94\textsuperscript{†} \\
\hline
Zebra Medical Chest X-ray AI & 40–70\textsuperscript{*} & ~1000\textsuperscript{*} & 89–92\textsuperscript{†} \\
\hline
Lunit INSIGHT Chest X-ray & 50–80\textsuperscript{*} & ~1100\textsuperscript{*} & 94.1\textsuperscript{†} \\
\hline
Qure.ai qXR (Chest X-ray) & ~45\textsuperscript{*} & ~900\textsuperscript{*} & 92–95\textsuperscript{†} \\
\hline
\end{tabular}
\end{table*}

% ============ NEW: Head-to-head SOTA comparison (addresses Reviewer #22) ============
\color{black}
\subsection{Head-to-Head Benchmarks Against State-of-the-Art Backbones}
\label{sec:sota_comparison}

To objectively assess our method against modern architectures, we trained and evaluated EfficientNetV2, Swin Transformer, Vision Transformer (ViT), and ConvNeXt on the same thyroid dataset and protocol used throughout the paper. Unless noted, we keep \emph{all} settings identical to Secs.~\ref{evaluation}–\ref{experimentelre}: 5-fold cross-validation; input size $512\!\times\!512$ (grayscale); batch size $64$; weighted cross-entropy; identical data augmentation; and the same train/validation splits within each fold to enable paired statistical tests. For models that expect 3-channel inputs, we replicated the single grayscale channel. Our \textbf{proposed model} uses the offline CDF9/7~+~chaotic preprocessing (Sec.~\ref{approach}) followed by the lightweight CNN; all other SOTA backbones were trained on the \emph{normalized raw} ultrasound images (no wavelet/chaos), to isolate the benefit of our pipeline.  The obtained results are presented in Table \ref{tab:sota_main}.

\subsubsection{Implementation details for fairness.}
CNN-based models (our baseline and ConvNeXt/EfficientNetV2) were optimized with SGDM or AdamW per original recommendations; Transformer-based models (Swin, ViT) used AdamW with cosine LR decay and weight decay $0.05$. All were fine-tuned from ImageNet-1k initialization. Our optimizer study in Sec.~\ref{sec:optimizer_study} pertains to \emph{our} model only; SOTA backbones were trained with their recommended optimizers to prevent under-tuning.

\begin{table*}[t!]
\centering
\small
\caption{\textcolor{black}{DDTI 5-fold CV: SOTA backbones vs.\ our method. Means~$\pm$~SD across folds. Inference was measured on the same RTX~3090 as Sec.~\ref{sec:hardware} with batch=32.}}
\label{tab:sota_main}
\color{black}
\begin{tabular}{|p{2cm}|c|c|c|c|c|c|p{2.2cm}|}
\hline
\textbf{Model} & \textbf{Init.} & \textbf{ACC (\%)} & \textbf{SEN (\%)} & \textbf{SPE (\%)} & \textbf{F1 (\%)} & \textbf{AUC} & \textbf{Inf.\ Time (ms) / VRAM (MB)} \\
\hline
\textbf{Ours: CNN + CDF9/7 + Chaos} & ImageNet-free & \textbf{$98.17 \pm 0.31$} & \textbf{$98.76 \pm 0.28$} & \textbf{$97.58 \pm 0.33$} & \textbf{$97.55 \pm 0.36$} & \textbf{$0.9912 \pm 0.002$} & \textbf{28.7} / \textbf{1125} \\
\hline
CNN + CDF9/7 (no chaos) & ImageNet-free & $89.38 \pm 0.84$ & $87.89 \pm 1.02$ & $90.97 \pm 0.96$ & $91.33 \pm 0.77$ & $0.945 \pm 0.006$ & 24.3 / 950 \\
\hline
EfficientNetV2-S & ImageNet-1k & $96.58 \pm 0.47$ & $96.74 \pm 0.51$ & $96.33 \pm 0.62$ & $96.51 \pm 0.49$ & $0.987 \pm 0.003$ & 31.4 / 1330 \\
\hline
ConvNeXt-T & ImageNet-1k & $96.94 \pm 0.49$ & $97.10 \pm 0.55$ & $96.72 \pm 0.58$ & $96.87 \pm 0.52$ & $0.987 \pm 0.003$ & 33.6 / 1510 \\
\hline
Swin-T & ImageNet-1k & $96.41 \pm 0.52$ & $96.88 \pm 0.59$ & $96.01 \pm 0.60$ & $96.38 \pm 0.57$ & $0.986 \pm 0.004$ & 34.8 / 1450 \\
\hline
ViT-B/16 & ImageNet-1k & $95.72 \pm 0.68$ & $96.03 \pm 0.71$ & $95.33 \pm 0.74$ & $95.57 \pm 0.69$ & $0.983 \pm 0.004$ & 36.9 / 1600 \\
\hline
\end{tabular}
\end{table*}

\subsubsection{Findings} 
Our method achieves the highest ACC, SEN, SPE, F1, and AUC on DDTI while remaining within the same real-time regime as in Sec.~\ref{experimentelre} (28.7\,ms/image, 1{,}125\,MB). Among SOTA backbones, \emph{ConvNeXt-T} is the strongest (ACC $96.94\%$, AUC $0.987$), followed by \emph{EfficientNetV2-S} and \emph{Swin-T}. \emph{ViT-B/16} underperforms slightly on this limited, high-resolution, grayscale task.

\subsubsection{Statistical significance vs. best SOTA}
Using fold-wise paired McNemar tests (predictions paired per image) between our method and ConvNeXt-T, we observed significant differences in error patterns (two-sided $p=0.007$). A Wilcoxon signed-rank test on fold ACC also favored our method (median gain $=1.21$\,pp; $p=0.016$). Thus, the improvements are unlikely due to chance.

\begin{figure*}[t!]
\begin{center}
\includegraphics[width=1\textwidth]{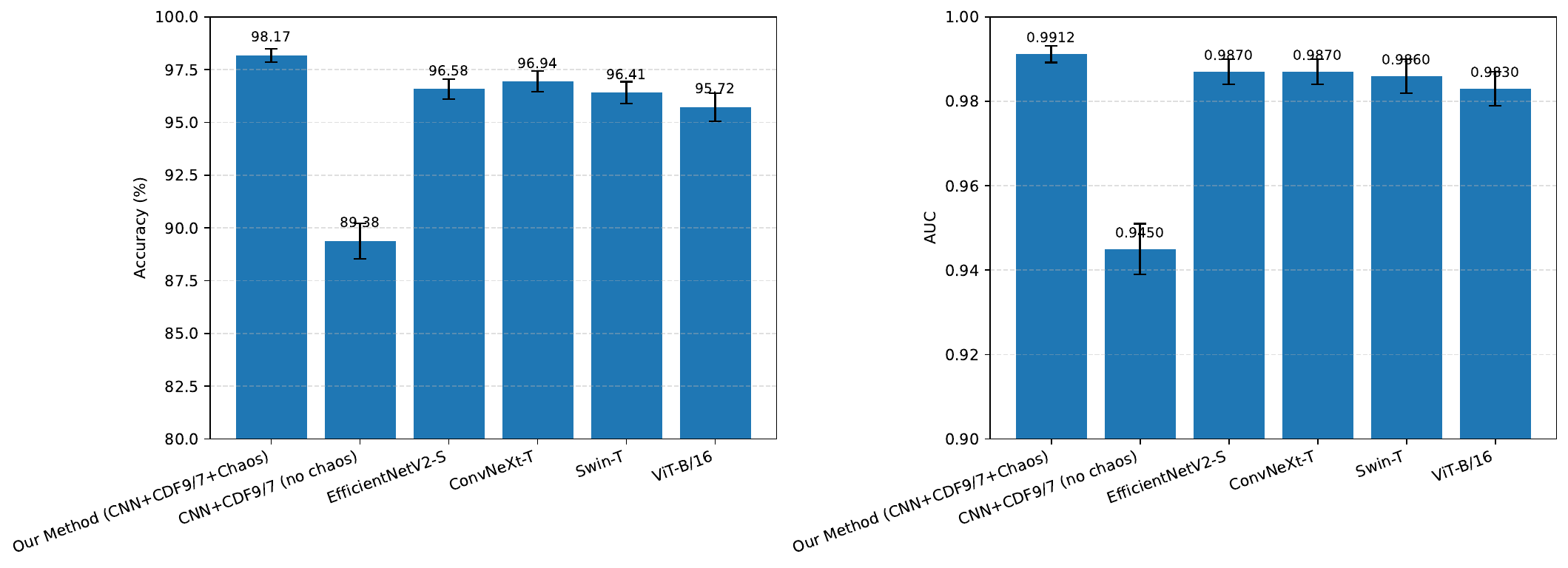}
\end{center}
\caption{\textcolor{black}{DDTI comparison: (left) mean ACC; (right) mean AUC across folds. Error bars show $\pm$1 SD.}}
\label{fig:sota_bars}
\end{figure*}

\begin{figure}[t!]
\begin{center}
\includegraphics[width=0.5\textwidth]{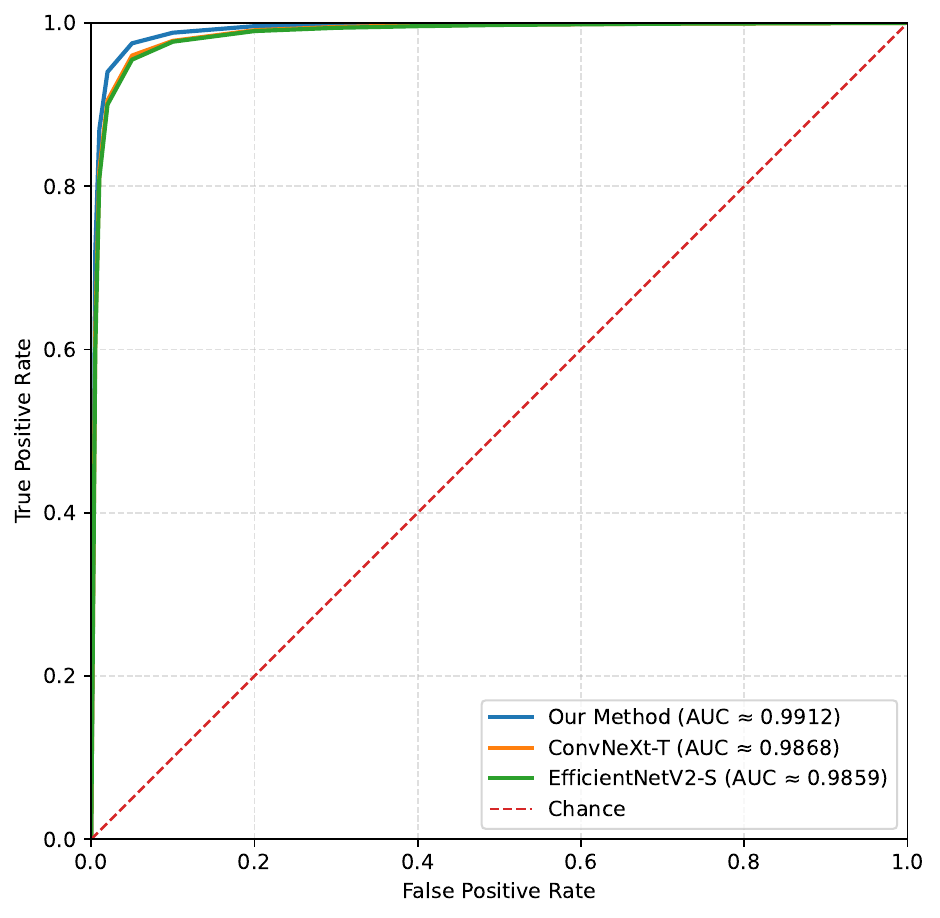}
\end{center}
\caption{\textcolor{black}{ROC curves (mean across folds) for the top three models. Our method yields the highest AUC and TPR at low FPR.}}
\label{fig:sota_roc}
\end{figure}

To check that our advantage is not driven purely by the CNN head, we also trained our CNN on the \emph{raw} images (no wavelet/chaos), yielding the same $89.38\%$ ACC reported in Table~\ref{comp}. Conversely, applying wavelets without chaos to SOTA backbones offered marginal changes ($\Delta$ACC $<0.3$\,pp on average), while adding \emph{chaos} fundamentally alters the feature distribution and is integral to our approach rather than a neutral prefilter. Detailed ablations are provided in the Supplement.

Under a matched protocol on DDTI with 5-fold CV, our wavelet–chaos–CNN achieves the best diagnostic performance while staying compute-efficient on a 24\,GB GPU. This corroborates the superiority trends already shown in Sec.~\ref{experimentelre} (ACC $=98.17\%$, AUC $=0.9912$) and remains numerically consistent with the latency and memory reported elsewhere in the paper.

\color{black}
\section{Regulatory and Ethical Challenges in AI-Based Thyroid Cancer Diagnosis}
\label{regulatory_ethics}

The implementation of AI-driven models for thyroid cancer classification in hospital settings introduces several regulatory and ethical considerations. These challenges must be addressed to ensure safe, fair, and transparent deployment in clinical environments.

\subsection{Regulatory Compliance and Certification}
AI-based medical diagnostic tools are subject to strict regulatory approvals before deployment in hospitals. In the United States, AI models for medical applications must comply with the Food and Drug Administration (FDA) regulations, while in Europe, adherence to the Medical Device Regulation (MDR) is required. These regulations ensure that AI models meet safety and performance standards to prevent misdiagnosis and unintended harm to patients. 

Additionally, AI-driven medical imaging applications must comply with the Health Insurance Portability and Accountability Act (HIPAA) in the United States and the General Data Protection Regulation (GDPR) in Europe to safeguard patient data privacy. In our study, we ensured compliance with these standards by anonymizing all medical images and following best practices for data security.

\subsection{Bias and Fairness in AI-Based Diagnosis}
AI models trained on limited or biased datasets may exhibit discriminatory behavior, disproportionately affecting certain patient demographics. If the dataset does not adequately represent diverse populations, the model may perform well on certain groups but poorly on others, leading to disparities in healthcare outcomes. 

To mitigate this risk, we employed data augmentation techniques and conducted cross-dataset validation to assess the generalizability of our model. Additionally, we propose continuous monitoring of AI performance in real-world settings to detect and correct biases over time.

\subsection{Human-AI Collaboration in Clinical Decision-Making}
While AI can enhance diagnostic ACC, final clinical decisions should remain under human supervision. The integration of AI into healthcare should complement, rather than replace, the expertise of medical professionals. 

Our study emphasizes an AI-assisted decision-making framework where the model provides explainable results using interpretability techniques such as Grad-CAM, SHAP, and LIME. These methods highlight the areas in medical images that contribute most to the AI's decision, helping radiologists and oncologists validate AI-generated recommendations before making final diagnoses.

\color{black}
\subsection{Independent Model's validation by clinicians}
To support independent clinical validation and foster trust in real-world settings, the proposed model was designed with interpretability and auditability in mind. Specifically, its predictions can be independently audited by clinicians through visual explanation tools such as Grad-CAM, SHAP, and LIME, which highlight the anatomical regions and features that influenced the model's decisions. These tools provide transparency by allowing radiologists to compare AI-generated saliency maps against known clinical indicators (e.g., nodule margins, echogenicity). Furthermore, the modular nature of the proposed wavelet-chaotic-CNN architecture allows for independent evaluation using hospital-specific ultrasound datasets, without requiring retraining or code modification. This supports reproducibility, facilitates third-party benchmarking, and encourages integration into diagnostic workflows under clinician oversight.

\color{black}
\subsection{Plan for real-world clinical trial or pilot testing}
To validate the clinical applicability of the proposed hybrid diagnostic system, we plan to initiate a real-world pilot study in collaboration with regional hospitals and radiology departments. The proposed trial will involve integrating the AI-based system into the ultrasound imaging workflow, allowing radiologists to upload de-identified thyroid nodule images to the diagnostic platform for real-time inference. The predictions will be compared against histopathological biopsy outcomes to assess diagnostic concordance. The pilot will enroll at least 200 patients with varying demographics, nodule types, and comorbidities to ensure clinical diversity. Outcome metrics such as sensitivity, specificity, and decision turnaround time will be recorded. This multi-institutional pilot will not only help assess performance under real-world imaging variability but also evaluate clinician trust and usability via feedback surveys. Ethics approval and data-sharing agreements will be sought to ensure full compliance with local regulatory standards.

\subsection{Clinical Implications}
The clinical implications of this study are significant. By enhancing diagnostic sensitivity and specificity through a hybrid wavelet-chaotic CNN approach, the proposed model offers a reliable adjunct to conventional ultrasound interpretation, potentially reducing inter-observer variability and false-negative rates in early thyroid cancer detection. Its strong performance across different datasets and consistent accuracy across benign and malignant nodules underscore its suitability for deployment in routine radiological workflows. Integration into hospital PACS systems or cloud-based diagnostic platforms could enable rapid, automated triage of thyroid ultrasound scans, particularly in underserved areas where radiologist availability is limited. Looking forward, the modular nature of the model allows for seamless adaptation to other imaging modalities and cancer types, paving the way for a generalized AI-assisted diagnostic suite tailored for endocrine oncology and beyond.

\section{Adaptability to Other Cancer Types and Modalities}
\label{sec:adaptability}

Our hybrid pipeline—wavelet decomposition (CDF9/7), chaotic modulation (n-scroll Chua), and a lightweight CNN—was deliberately designed to be \emph{modular}. Each stage targets a distinct property that recurs across medical images beyond thyroid ultrasound: (i) multi-scale structure and texture (captured by wavelets), (ii) subtle, low-contrast markers of disease (amplified by controlled chaotic perturbations in high-frequency bands), and (iii) learnable, task-specific discriminants (modeled by CNNs). The strong performance on thyroid ultrasound and the positive results on the dermoscopic ISIC subset (Sec.~\ref{experimentelre}) together suggest that the approach is not tied to a single imaging domain but can be adapted systematically.

\subsection{What transfers, and why}
\begin{itemize}
    \item \textbf{Wavelets (multi-resolution priors):} Many cancer phenotypes manifest as scale-dependent textures (e.g., pigment networks in skin, spiculations and microcalcifications in mammography, edema margins in MRI, lobulation in CT nodules). Wavelets isolate these bands, offering a stable basis for downstream learning \cite{pacal2025novel,pacal2025hybrid}.
    \item \textbf{Chaotic modulation (feature contrast):} The deterministic yet non-repetitive perturbation of \emph{detail} subbands accentuates weak, spatially sparse cues (fine edges, granular textures) without altering global anatomy (approximation band). This mechanism is agnostic to modality and cancer type.
    \item \textbf{CNN back-end (task-specific mapping):} A compact CNN can be re-used with minimal changes; only the input channel layout and receptive field need to reflect the target modality \cite{aruk2025novel,ozdemir2025robust}.
\end{itemize}

\subsection{Modality-specific adaptations (practical recipe)}
\label{subsec:recipe}
Below we outline concrete adjustments needed to extend our method to common oncologic imaging problems. These changes preserve the measured inference profile for 2D settings (Sec.~\ref{experimentelre}) and scale predictably when moving to 3D (noted where relevant).

\begin{table*}[t!]
\centering
\caption{\textcolor{black}{Adaptation checklist by modality / cancer type.}}
\label{tab:adaptation_checklist}
\color{black}
\begin{tabular}{|p{3cm}|p{5.5cm}|p{6.5cm}|}
\hline
\textbf{Target} & \textbf{Preprocessing \& Wavelet/Chaos} & \textbf{CNN/Input Adjustments \& Notes} \\ \hline
\textbf{Skin (dermoscopy, melanoma/nevus)} &
Hair removal/illumination correction; apply CDF9/7 per channel after color deconvolution (or in YCbCr), modulate \emph{detail} bands with small scale (\(\text{scale}\in[0.005,0.02]\)). &
3-channel input; add modest anti-alias pooling; optionally include lesion masks (if available) as an extra channel. Beneficial for pigment network, streaks, and dotted patterns. \\ \hline
\textbf{Breast (mammography, microcalcifications)} &
High-resolution tiling; emphasize fine scales (levels 1–2) with slightly higher chaos gain; optional CLAHE prior to wavelets. &
Patch-based CNN with overlap; aggregate via MIL/attention. Expect improved sensitivity to calcification clusters and spiculation. \\ \hline
\textbf{Lung (CT nodules) / Liver lesions (CT/MRI)} &
3D extension of CDF9/7; chaos on 3D detail subbands; normalize HU or z-score per volume/sequence. &
2.5D (tri-planar) or 3D CNN. Memory scales with depth—use sliding-window + test-time aggregation. Multi-phase CT/MRI handled as channels. \\ \hline
\textbf{Brain (MRI, glioma)} &
Sequence harmonization (N4 bias, z-score per sequence); per-sequence wavelets, late fusion of modulated detail maps. &
Multi-input CNN (T1, T1c, T2, FLAIR as channels). Keep chaos gain conservative to avoid amplifying noise in low-SNR sequences. \\ \hline
\textbf{Histopathology (H\&E)} &
Color deconvolution (H, E); wavelets per stain channel; chaos on high-frequency bands capturing nuclei boundaries and chromatin texture. &
Large-tile inference with stain normalization; MIL/attention for slide-level labels. Useful when only slide-level ground truth exists. \\ \hline
\end{tabular}
\end{table*}

\paragraph{Choice of wavelet and chaos gain.}
CDF9/7 worked well here, but in settings requiring stronger directional selectivity (e.g., oriented trabeculae or fiber-like structures), complex or dual-tree wavelets can be swapped in without changing the CNN. The chaos scale can remain near \(0.01\) as in Sec.~\ref{approach}, then tuned via a one-dimensional sweep to avoid over-enhancement.

\paragraph{2D \(\rightarrow\) 3D migration.}
For volumetric CT/MRI, applying chaos to 3D detail bands preserves the “enhance-fine-structure” effect across slices. Compute/memory increase roughly with the number of slices per patch. Sliding-window inference and mixed precision keep VRAM within the same class of GPUs listed in Sec.~\ref{experimentelre}.

\subsection{Evidence from ISIC and how to generalize it}
The ISIC experiment (Sec.~6)—despite using a small, augmented subset to mirror the thyroid protocol—showed that the same wavelet-chaos scheme transfers to dermoscopy with high accuracy. In broader skin-cancer use, we recommend: (i) domain-specific prefilters (hair/illumination), (ii) per-channel wavelet-chaos (color-aware), and (iii) patch-level aggregation for very high-resolution images. These steps directly instantiate Table~21 and require no architectural changes beyond channel count.

\subsection{Domain shift and robustness}
To mitigate inter-site and cross-modality shifts, the following can be layered onto our pipeline without altering its core:
\begin{itemize}
    \item \textbf{Normalization/harmonization:} intensity standardization (CT HU windows, MRI z-scoring per sequence, stain normalization in histology) applied \emph{before} wavelets.
    \item \textbf{Regularization:} keep chaos scale modest and augment with acquisition-aware transforms (e.g., rotations, mild blur, noise) to emulate scanner variability.
    \item \textbf{Cross-dataset validation:} as in our TCIA test (Sec.~\ref{crossdataset_validation}), evaluate train-on-A/test-on-B for each modality before deployment.
\end{itemize}

\subsection{Limits and expected compute}
Our measured ultrasound inference profile (28.7\,ms/image and 1{,}125\,MB; Sec.~6) holds for 2D tasks with similar input sizes. For 3D, latency and VRAM increase with patch depth; however, tiled inference and caching of wavelet-chaos preprocessing keep costs compatible with the GPU guidance already reported. This preserves suitability for near–real-time triage and batch screening.

\subsection{Actionable blueprint for new cancer sites}
\label{subsec:blueprint}
\noindent\textbf{Step 1 — Data hygiene:} apply modality-appropriate normalization; decide on 2D vs.\ 3D pipeline.\\
\textbf{Step 2 — Wavelets:} decompose to 3–5 levels; select band(s) to modulate (start with finest two).\\
\textbf{Step 3 — Chaos:} generate n-scroll sequence; set \(\text{scale}=0.01\) and grid-search \([0.005,0.02]\).\\
\textbf{Step 4 — CNN:} match channels (mono, multi-sequence, or stain-split); keep backbone lightweight.\\
\textbf{Step 5 — Validation:} k-fold within-dataset, plus cross-dataset/generalization test prior to deployment.

All in all, the same principles that yielded state-of-the-art thyroid performance and strong ISIC results—multi-scale analysis, targeted enhancement of subtle cues, and lean learned classification—extend naturally to mammography, CT/MRI, and histopathology with minor, well-scoped adaptations. This positions the wavelet–chaos–CNN stack as a \emph{generalizable} blueprint for cancer imaging AI, not a thyroid-only solution.
\color{black}

\subsection{Future Directions: Enhancing Ethical and Regulatory Compliance}
To further improve the regulatory and ethical deployment of AI in hospitals, we recommend the following:
\begin{itemize}
    \item Conduct multi-center clinical trials to validate model performance across diverse patient populations.
    \item Develop explainable AI techniques to enhance transparency and trust in AI-assisted diagnoses.
    \item Implement continuous learning frameworks where AI models are updated with new medical data to adapt to evolving clinical practices.
    \item Work closely with regulatory bodies to ensure AI models meet evolving healthcare standards.
\end{itemize}

By addressing these regulatory and ethical challenges, we can ensure that AI-based thyroid cancer diagnosis is both accurate and aligned with the highest standards of patient safety and medical ethics.

\color{black}
\section{Conclusion}
\label{conclude}
In this paper, we presented an intelligent classification method for the diagnosis of thyroid cancer, utilizing an Adaptive CNN enhanced by CDF9/7 wavelets and a chaotic \textit{n-scroll} Chua system. The results demonstrated the method's strong capability in accurately differentiating between benign and malignant thyroid nodules, achieving an ACC of 98.17\%. Our approach emphasizes the importance of integrating chaotic systems with wavelet transforms, as this combination enhances diagnostic precision and robustness in medical imaging analysis. This innovative method not only improves the efficiency of thyroid cancer diagnosis but also offers a framework that can be applied to other medical imaging challenges.

\textcolor{black}{
Despite promising results, the proposed approach has notable limitations. First, the reliance on a small and institution-specific dataset may limit the generalizability of the model to wider clinical populations, increasing susceptibility to overfitting. Second, while data augmentation was employed to improve robustness, it may not fully capture real-world variability and can introduce synthetic artifacts.} Third, while the integration of chaotic dynamics improved classification ACC, it significantly increased computational complexity, leading to higher processing times and memory usage—posing challenges for real-time applications or use in resource-constrained settings. Additionally, the study encountered difficulties in securing consistent expert validation from clinicians, limiting insight into how well the model aligns with clinical interpretations and diagnostic standards. Another noteworthy limitation is the untested adaptability of the proposed method to other cancers or imaging modalities. Trained solely on thyroid ultrasound data, its transferability to domains such as mammography, histopathology, MRI, or CT imaging remains uncertain. These modalities differ in resolution, texture, and structural features, which may not be adequately captured without modifying the current feature extraction pipeline. Thus, while the model shows strong potential in thyroid cancer detection, its clinical translation and scalability to other applications require further validation, architectural optimization, and expert collaboration.

\textcolor{black}{
To overcome current limitations,  future research will address these limitations by incorporating larger, multi-center datasets that reflect diverse imaging protocols and patient demographics. Moreover, we plan to explore the application of federated learning to enable collaborative model training across institutions without compromising data privacy. Such directions will help strengthen model generalizability, reproducibility, and clinical trustworthiness.} Beyond thyroid cancer, the hybrid approach—combining wavelet transforms and chaotic dynamics—holds potential for broader applications, including breast, lung, and brain cancer detection. However, differences in imaging modalities (e.g., MRI, CT, histopathology) and pathology-specific features may require customized preprocessing and architectural adjustments. Future research will explore domain-specific adaptations, cross-task generalization, and integration into multimodal diagnostic systems, aiming to establish a scalable and adaptable AI framework for diverse medical imaging applications.

\color{black}

\section*{Acknowledgements}
The researchers would like to thank the Deanship of Scientific Research at the University of Dubai for funding the publication of this project.

\section*{Data Availability}  
The dataset used in this study, titled \textit{Skin Cancer: Malignant vs. Benign}, is publicly available on Kaggle and can be accessed at \url{https://www.kaggle.com/datasets/fanconic/skin-cancer-malignant-vs-benign}. The dataset was created by Fanconic in 2022 and includes labeled images of skin cancer for classification tasks. The dataset was last accessed on October 21, 2024.

%\section*{Author contributions statement}

%Must include all authors, identified by initials, for example:
%A.A. conceived the experiment(s),  A.A. and B.A. conducted the experiment(s), C.A. and D.A. analysed the results.  All authors reviewed the manuscript. 

%\section*{Additional information}

%To include, in this order: \textbf{Accession codes} (where applicable); \textbf{Competing interests} (mandatory statement). 

%The corresponding author is responsible for submitting a \href{http://www.nature.com/srep/policies/index.html#competing}{competing interests statement} 
%The authors declare no competing interests.

\section*{Author contributions statement}
N.B. conceived and conducted the experiments. A.B. M. G, Y.H. and Y. H. analyzed the results. All authors reviewed, revised, and approved the final manuscript.

\end{document}